\documentclass[11pt]{scrartcl}       
\usepackage[bottom=2.5cm,top=2.5cm,left=2.5cm,right=2.5cm]{geometry}

\usepackage{amsmath,amssymb,amsthm,graphicx,caption,cases}	
\usepackage[bottom]{footmisc}	
\usepackage{algorithmic}

\usepackage{abstract,amsfonts,amsbsy,amssymb,amstext,mathrsfs,latexsym,color,url,amsthm,xcolor,authblk,fontenc,bm}

\definecolor{NUSBlue}{RGB}{0,61,124} 
\definecolor{NUSOrange}{RGB}{239,124,0}
\usepackage[colorlinks=false,allbordercolors=NUSOrange]{hyperref}

\DeclareOldFontCommand{\bf}{\normalfont\bfseries}{\mathbf}

\DeclareMathOperator{\diag}{diag}

\newcommand{\bV}{\mathbf{V}}
\newcommand{\bI}{\mathbf{I}}

\newcommand{\bB}{\mathbf{B}}
\newcommand{\bD}{\mathbf{D}}
\newcommand{\bL}{\mathbf{L}}
\newcommand{\bP}{\mathbf{P}}
\newcommand{\bQ}{\mathbf{Q}}
\newcommand{\bS}{\mathbf{S}}
\newcommand{\bC}{\mathbf{C}}
\newcommand{\bU}{\mathbf{U}}
\newcommand{\bE}{\mathbf{E}}

\newcommand{\bG}{\mathbf{G}}
\newcommand{\bW}{\mathbf{W}}
\newcommand{\bX}{\mathbf{X}}
\newcommand{\bY}{\mathbf{Y}}
\newcommand{\bZ}{\mathbf{Z}}

\newcommand{\bz}{\mathbf{z}}

\newcommand{\bv}{\mathbf{v}}

\newcommand{\ba}{\mathbf{a}}
\newcommand{\bb}{\mathbf{b}}
\newcommand{\bc}{\mathbf{c}}
\newcommand{\bd}{\mathbf{d}}
\newcommand{\be}{\mathbf{e}}
\newcommand{\bu}{\mathbf{u}}

\newcommand{\bp}{\mathbf{p}}
\newcommand{\bq}{\mathbf{q}}
\newcommand{\bA}{\mathbf{A}}

\newcommand{\bx}{\mathbf{x}}
\newcommand{\by}{\mathbf{y}}

\newcommand{\cA}{\mathcal{A}}
\newcommand{\cB}{\mathcal{B}}
\newcommand{\cP}{\mathcal{P}}

\newcommand{\R}{\mathbb R}

\newcommand{\N}{\mathbb N}

\definecolor{NUSBlue}{RGB}{0,61,124}   

\usepackage{tikz}
\usepackage{mathtools}

\def\nudge{.5}

\tikzset{axis/.style={ultra thin, Grey, -latex, shorten <=-\nudge cm, shorten >=-2*\nudge cm}}
\tikzset{line/.style={thick}}

\usepackage{textcomp}
\DeclareMathAlphabet\mathbfcal{OMS}{cmsy}{b}{n}
\usepackage{hyperref}
\usepackage[linesnumbered,vlined,ruled]{algorithm2e}
\usepackage{amsmath,amsfonts,amssymb,amsthm}
\theoremstyle{plain}
\newtheorem{thm}{Theorem}

\newtheorem{exam}[thm]{Example}

\newtheoremstyle{cited}%
  {3pt}
  {3pt}
{\itshape}
  {}
  {\bfseries}
  {.}
  {.5em}
  {\thmname{#1} \thmnumber{#2} \thmnote{\normalfont#3}}
\theoremstyle{cited}
\newtheorem{citedthm}[thm]{Theorem}
\newtheorem{citedlem}[thm]{Lemma}
\newtheorem{citedcor}[thm]{Corollary}
\newtheorem{citeddef}[thm]{Definition}
\newtheorem{citedprop}[thm]{Proposition}

\usepackage{graphicx}
\usepackage{epstopdf}
\usepackage{epsf}
\usepackage{subfigure}
\usepackage{epsfig}

\usepackage{cite}

\usepackage{tabularx}
\usepackage{array}
\usepackage{booktabs}

\usepackage{comment}
\usepackage{tgpagella} 
\usepackage{mathpazo}

\begin{document}
\renewcommand*{\Authsep}{, }
\renewcommand*{\Authand}{, }
\renewcommand*{\Authands}{, }
\renewcommand*{\Affilfont}{\normalsize}   
\setlength{\affilsep}{2em}   
\title{Representation and decomposition of functions in DAG-DNNs and structural network pruning}
\date{\today}
\author{Wen-Liang Hwang}
\maketitle

\begin{abstract}
The conclusions provided by deep neural networks (DNNs) must be carefully scrutinized to determine 
whether they are universal or architecture dependent. 
The term DAG-DNN refers to a graphical representation of a DNN in which the architecture is expressed as a direct-acyclic graph (DAG), on which arcs are associated with functions. The level of a node denotes the maximum number of hops between the input node and the node of interest. In the current study, we demonstrate that DAG-DNNs can be used to derive all functions defined on various sub-architectures of the DNN. We also demonstrate that the functions defined in a DAG-DNN can be derived via a sequence of lower-triangular matrices, each of which provides the transition of functions defined in sub-graphs up to nodes at a specified level. The lifting structure associated with lower-triangular matrices makes it possible to perform the structural pruning of a network in a systematic manner. The fact that decomposition is universally applicable to all DNNs means that network pruning could theoretically be applied to any DNN, regardless of the underlying architecture. We demonstrate that it is possible to obtain the winning ticket (sub-network and initialization) for a weak version of the lottery ticket hypothesis, based on the fact that the sub-network with initialization can achieve training performance on par with that of the original network using the same number of iterations or fewer.
\end{abstract}
%

\section{Introduction}

The deep neural networks (DNNs) used for classification and other tasks continue to astound researchers with their prowess. Despite theoretical findings involving networks of infinite-width or infinite-depth, such as the consistency in classification \cite{radhakrishnan2023wide}, the universal approximation theory \cite{Cybenko1989,Hornik91,heinecke2020refinement}, and the connections to Gaussian processes \cite{neal2012bayesian,williams1998prediction,lee2017deep} and neural tangent kernels \cite{jacot2018neural}, 
our understanding of the theoretical underpinnings is still far from complete.  Our lack of understanding can be attributed at least in part to a lack of tools by which to analyze the composition of non-linear activation functions in DNNs and a lack of mathematical models adaptable to a diversity of DNN architectures.
If we disregard the output layer, then the finite-depth DNNs commonly used for theoretical analysis can be viewed as a sequence of layers, wherein a layer denotes a non-linear activation function followed by an affine linear mapping, as follows: $ M_L \circ \varrho_{L-1} \circ\cdots\circ\varrho_1\circ M_1(\bx)$.
The network above comprises $L\in\N$ layers with a width of $\{N_{\ell}\in\N\}$,  a collection of affine linear operators $\{M_{\ell}\colon \R^{N_{\ell-1}}\to\R^{N_{\ell}} \}_{\ell=1}^L$, and non-linear activation functions $\{\varrho_{\ell}\}_{\ell=1}^{L-1}$. The theoretical results derived using this architecture no doubt capture some intrinsic properties of DNNs; however, the architecture is not sufficiently general to preclude the possibility that some of the results are architecture-dependent. One example is the inclusion of additional network layers, which are believed to refine partitions within the input space in the form of tree-like partitions of polytopes \cite{BaraniukPowerDiagramSubdiv, hwang2019rectifying}. The input space of a DNN is partitioned in a coarse-to-fine manner for function approximation, but not generally in a tree-like partition due to the use of concatenation (fusion) operations combining outputs from several channels \cite{hwang2022analysis}.

The term ``DAG-DNN" refers to a graphical representation of a DNN, in which the architecture is expressed as a direct-acyclic graph (DAG) for use in exploring the functional properties of a DNN, such as input domain partitioning and the asymptotic stability of a function with respect to small input perturbations.
Note that the axiomatic approach to constructing DAG-DNNs is applicable to a wide range of  DNN architectures \cite{hwang2022analysis}. A DAG-DNN is recursively constructed by applying a single regulatory rule to axiomatic operations associated with basic elements, which means that the network can no longer be delineated based on ``layers". Rather, it is preferable to delineate the network based on the notion of ``levels"; i.e., a collection of nodes that shares the same maximum number of hops to the input node. In other words, nodes can be partitioned according to their levels. Further a function can be associated to a node defined on the sub-graph proceeding from the input node to the node of interest. 
The fact is that the association between a sub-graph and function can be extended to every pair of nodes only if the function from one node to the other node is defined. 
In this paper, we demonstrate the means by which a DNN can be transformed into an equivalent DAG-DNN and the derivation of functions for every pair of nodes within it.
We demonstrate that the matrix recording all-pair functions can be decomposed based on node level into a sequence of lower triangular matrices, each of which characterizes the transformation of functions in every sub-graphs covering nodes up to a given level into functions of sub-graphs of nodes ending at the next level. 
This is referred to as the lifting structure, describing how the complexity of a function is lifted from one level to the next. Note that function approximations that use the lifting modulus for decomposition are universal applicable to all DAG-DNNs.

This paper also presents conditions for the structural pruning of a network based on the lifting modulus. We show that sub-networks with initializations obtained via structural pruning can theoretically achieve training performance on par with that of the original network. 
This analytical result can be viewed as a weak version of the lotus ticket hypothesis \cite{frankle2018lottery}, wherein the testing performance of a pruned sub-network based on empirical findings can generally be as good as that of the original network as long as the sub-network is properly initialized. This hypothesis is supported in theory when applied to over-parameterized two-layer neural networks (with one hidden-layer of neurons)  based on weight pruning (some weight coefficients are set to zero) \cite{malach2020proving} and structural pruning (some neurons are removed) \cite{zhang2021lottery}. Note however that the results cannot be extended to the highly complex DNNs typically encountered in the real world. 
Despite the fact that our theory is applicable only to training performance, it can still be used for structural pruning of any DAG-DNN, regardless of the underlying structure.

The remainder of the paper is organized as follows.  
In Section~\ref{sec:DAGsec}, we review connected DAG-DNNs and previous works related to network compression. Section~\ref{matrixcomp} outlines the matrix representation and algebra of functions defined for a DAG-DNN. Section~\ref{sec:representation} presents a representation of functions defined for every sub-graph in a DAG-DNN.  The representation can be factored into multiplications of lower-triangle matrices, demonstrating progressively function transitions in accordance with levels of the nodes. Section~\ref{sec:compression} demonstrates that structural pruning can be universally applied to obtain sub-networks with training performance as good as that of the original network. Concluding remarks are presented in Section \ref{sec:conclusions}.

\textbf{Notation:} \\
Matrices are denoted using bold upper case letters and vectors are denoted using bold lower case letters. In graphs, a solid circle indicates a concatenation node. Double-circled nodes in figures denote level-domain nodes, which can be used to denote either the collection of nodes at a given level or those leading up to a given level, depending on the context.

\section{Related works}\label{sec:DAGsec}

\subsection{Connected DAG-DNNs} 

A DAG-DNN is a representation of a DNN in the form of a directed acyclic graph (DAG) in accordance with axiomatic rules. Arcs are attached using functions, while nodes relay or reshape vector dimensions to match the input/output of a function.
DAG-DNNs are defined by activation functions ($\rho$s), non-linear transformations ($\sigma$s), and underlying axiomatic rules (O1-O3). 
As demonstrated in \cite{hwang2022analysis}, DAG-DNNs are analyzable and generalizable to cover all DNNs in common use. Activation functions and non-linear transformations are both functions; however, un-rectifying analysis treats them differently \cite{hwang2019rectifying}. Activation functions are referred to as continuous piece-wise (CPWL) functions, where data-dependence can be explicitly depicted in the representation of a function. ReLU and max-pooling are activation functions, whereas the sigmoid and soft-max are viewed as non-linear transformations.
The activation functions of DAG-DNNs are assumed to be expressible as networks of point-wise CPWL activation functions comprising a finite number of pieces. The following lemma asserts that the activation functions of concern are a linear composition of translated-ReLUs.

\begin{citedlem}\cite{arora2016understanding} \label{RELU1}
Any point-wise CPWL activation function $\rho: \R \rightarrow \R$ of $m$ pieces can be expressed as follows:
\begin{align} \label{PWm}
\rho(x) 
& = \sum_{i=1}^m r_i \text{ReLU}(x - a_i) + l_i \text{ReLU}(t_i - x) \nonumber\\
& = \sum_{i \in I^+} r_i \text{ReLU}(x - a_i) +  \sum_{i \in I^-} l_i \text{ReLU}(t_i - x)
\end{align}
where $l_i$ and $r_i$ indicate the slopes of segments, and $a_i$ and $t_i$ are the breakpoints of the corresponding segments. 
\end{citedlem}
To enhance the stability of a very deep DAG-DNN against input perturbations, additional assumptions can be imposed on $\rho$s and $\sigma$s \cite{hwang2022analysis} to achieve bounded Lipschitz regularity, regardless of the number of DNN layers.

\begin{figure}[th]
\centering
\subfigure[A series-connection (O1)]{%
\includegraphics[width=0.5\textwidth]{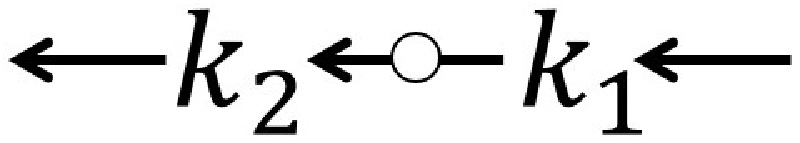}} \\
\subfigure[Concatenation (O2)]{\includegraphics[width=0.25\textwidth]{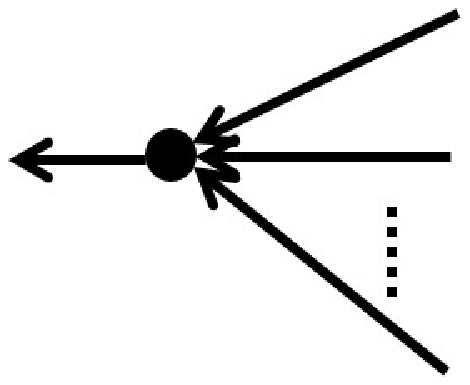}}
\subfigure[Duplication (O3)]{\includegraphics[width=0.25\textwidth]{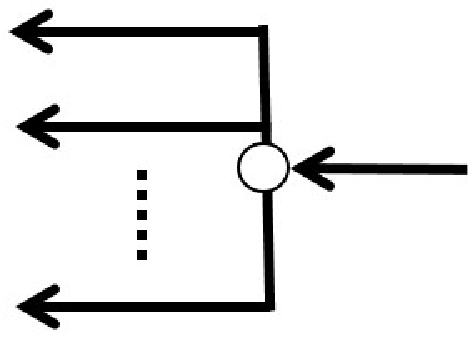}} 
\caption{Graphical representation of atomic operations, where functions attached to arcs of concatenation and duplication are referred to as identify functions $\bI$ (omitted for brevity) and the solid node in (b) differentiates the corresponding operation. Note that a reshaping of input or output vectors is applied at nodes to validate these operations.} 
\label{figbasic}
\end{figure}

DAG-DNNs are constructed by applying a sequence of three atomic operations (O1-O3) on functions in the basic set in accordance with the regulatory rule (R), which describes a legitimate method by which to apply an atomic operation to DAG-DNNs to yield another DAG-DNN. Basis set $\cB$ comprises functions of activations as well as linear and non-linear transformations, as follows:
\begin{align} \label{B1}
\cB = \{ \bI, \bL, M, \Gamma_{\rho}, \rho M, \Gamma_{\sigma}, \sigma M \},
\end{align}
where $\bI$ denotes the identify function; $\bL$ denotes any finite dimensional linear mapping with a bounded spectral-norm; $M = (\bL, \bb)$ denotes any affine linear mapping where $\bL$ and $\bb$ respectively refer to the linear and bias terms; $\Gamma_{\rho}$ denotes activation functions; $\rho M$ denotes functions with $\rho \in \Gamma_{\rho}$; $\Gamma_{\sigma}$ denotes non-linear transformations; and $\sigma M$ denotes functions with $\sigma \in \Gamma_{\sigma}$.
We denote $\chi$ as the input space for any elements in $\cB$. The corresponding DAG representations for O1-O3 are presented in Fig. \ref{figbasic}.
\begin{enumerate}
\item[O1.] Series-connection ($\circ$): Composition in which the output of  DAG-DNN $k_1$ is the input of function $k_2  \in \cB$, where
\begin{align*}\cB \circ \text{DAG-DNN}: k_2 \circ k_1:\bx \rightarrow k_2(k_1 \bx). \end{align*}
\item[O2.] Concatenation: Merging of multi-channel inputs $\bx_i \in \chi$ into a vector,  as follows:
\begin{align*}
\text{concatenation}: \{\bx_1, \cdots, \bx_m\} \rightarrow [\bx_1^\top \; \cdots \bx_m^\top]^\top.
\end{align*}
\item[O3.] Duplication: Duplication of an input to generate $m$ copies of itself, as follows: 
\begin{align*}
\text{duplication}: \bx \in \chi \rightarrow \begin{bmatrix}
\bI \\
\vdots \\
\bI 
\end{bmatrix} \bx.
\end{align*}
\end{enumerate}

Regulatory rule R generates other DAG-DNNs by regulating the application of atomic operations O1-O3 on DAG-DNNs. This rule precludes the generation of graphs that contain loops.

\begin{figure}[h]
\centering
{
\subfigure[]
{\includegraphics[width=0.2\columnwidth]{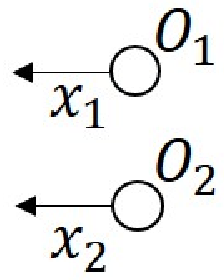}}
\subfigure[]{
\includegraphics[width=0.3\columnwidth]{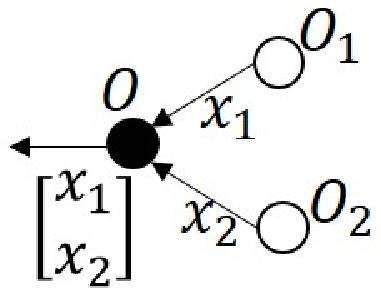}} \\
\subfigure[]
{\includegraphics[width=0.17\columnwidth]{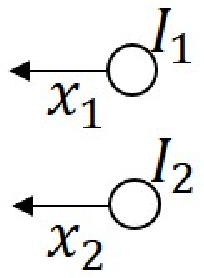}}
\subfigure[]{
\includegraphics[width=0.4\columnwidth]{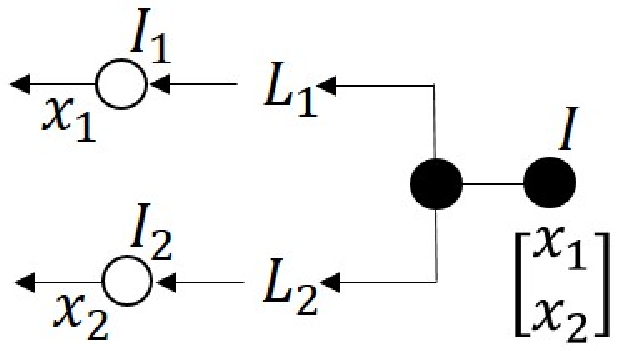}}}
\caption{(a), (b): Output nodes $O_1$ and $O_2$ in network (a) are concatenated to yield equivalent network (b) with a single output node; (c), (d): Input nodes $I_1$ and $I_2$ in network (c) are concatenated,  duplicated, and then selected using $L_1$ or $L_2$ in a separate channel to yield equivalent network (d) with one input node. 
} \label{IO}
\end{figure}

A DAG-DNN comprises nodes (vertices) and arcs generated by operations initiated by O1-O3 and R. As shown in Fig. \ref{IO}, we can assume that a DAG-DNN has only one input node $I$ and one output node $O$. It is deemed  to be connected if every vertex in the graph is reachable by the input node. In this paper, we consider only connected DAG-DNNs (simply referred to as DAG-DNNs for brevity), unless otherwise specified.  
The retention of DAG-DNNs after undergoing an operation is crucial to our analysis, as this allows the ordering of nodes in the DAG-DNN. Hereafter, we refer to simple figures to facilitate an understanding of DAG-DNNs.

\subsection{Network pruning}

Neural network pruning technique is used to to obtain a sparse, trainable sub-network with performance comparable to the original network or even better. 
This method is often used to obtain a favourable trade-off between sizes and function approximations when deploying large-scale network on smaller devices (e.g., mobile phones). A typical pruning algorithm follows the one-hop pipe-line used in the training of a large-scale network by which redundant parameters are pruned to obtain a sub-network, and the remaining parameters of which are fine-tuned. Considerable research effort has gone into the removal of un-necessary weights and structure (neurons or channels) from the original network. Empirical findings have been summarized as the lottery ticket hypothesis \cite{frankle2018lottery}. which states that a randomly-initialized dense neural network contains a sub-network that is initialized such that - when trained in isolation - it can match the test accuracy of the original network after undergoing the same number of training iterations or fewer. The sub-network with the initialization is referred to the winning ticket of the original network. Note that this hypothesis subsumes the pipe-line approach as a special case.

This hypothesis has inspired the development of advanced methods beyond the one-hop pipe-line approach. 
As experimentally demonstrated in \cite{frankle2018lottery}, it is possible to eliminate more than $90 \%$ of the parameters without compromising accuracy simply by applying multiple pruning loops, rewinding the weight initialization of the remaining coefficients, and re-training the sub-network. A sequence of techniques by which to capture winning tickets that comfort to the hypothesis are proposed: 1) rewind weight initializations and learning rates in stochastic gradient descents (SGDs) \cite{renda2020comparing}, 2) retain the signs of un-masked weight coefficients in re-training, and 3) investigate mask conditions to facilitate re-training \cite{zhou2019deconstructing}. 


The lottery ticket hypothesis can theoretically be justified \cite{malach2020proving, zhang2021lottery} by leveraging the redundancy in over-parameterized networks (where the number of parameters exceeds that of training data). 
In \cite{zhang2021lottery}, the authors describe the use of a two-layer oracle network for data generation. The goal is to simulate/recover the oracle using a trained two-layer network, which utilizes the data generated by the oracle as its training data. The trained model has the same number of neurons but smaller number of weight coefficients as the oracle (corresponding to weight pruning). Their analysis focused on harnessing the error between the weight coefficients of the oracle and the trained model versus the amount of training data, thereby establishing a link between sample complexity and the hypothesis. In \cite{malach2020proving}, 
probability is used to measure the existence of a sub-network satisfying the hypothesis within over-parameterized neural networks. These networks consist of either a two-layer structure for structural pruning or network architecture $M_L \circ \varrho_{L-1} \circ\cdots\circ\varrho_1\circ M_1(\bx)$ for weight pruning. 
Their analysis demonstrated that such sub-networks can achieve accuracy levels similar to those of the original network with a high probability without the need for training. Note however that deriving a suitable sub-network based on the above analysis is difficult in practical situations.



\section{Representation of DAG-DNNs using matrix compositions} \label{matrixcomp}

While representing networks using DAGs makes explicit the structure of the networks,  when considering the properties of network functions, representation a network as matrix compositions can be facilitated by leveraging the matrix algebra in function calculations.

Here and after, DAG-DNNs are referred to DNNs that can be represented as connected DAGs, in which each arcs is associated with an atomic operation, to which are attached function in the base set (\ref{B1}). 
The simplest DAG-DNN comprises only one node. Nodes and arcs are recursively attached to a DAG-DNN by  applying an atomic operation in accordance with the regulatory rule by which to yield another DAG-DNN.
The adjacent matrix $\bG$ of a DAG-DNN is a matrix in which the rows and columns are nodes and the value of an entry is either $1$ or $0$, in accordance with the following:
\begin{align}
\bG(i,j) = \begin{cases}
1,  \text{ there is an arc from node $j$ to node $i$;} \\
0, \text{ if $i = j$;} \\
0, \text{ otherwise}.
\end{cases}
\end{align}
Transitive closure is the term used to indicate whether one node is accessible to another node in a graph.
$\bG^k$ (the matrix derived by multiplying $\bG$ to itself $k$ times) displays transitivity closure, such that $\bG^k(i,j)$ gives a number of paths with a length precisely equal to $k$ from node $j$ to node $i$ in the graph. 


One of the objectives in this paper is to derive a matrix representation for each DAG-DNN, such that every entry of the matrix denotes the function of a sub-graph with input at one node and output at another node. 
We shall see that the functions on entries of the matrix can be deduced via a sequence of $\tilde \cB$-matrix multiplications (in a manner similar to the transitivity closure of an adjacent matrix).

Denote that $\cB \subseteq \tilde \cB$, and $\tilde \cB$ is closure under $\circ$ (composition) and $+$ (addition) operations such as
\begin{align} \label{K1}
\tilde \cB = \cB \cup \{ \tilde \cB \circ \tilde \cB \} \cup \{ \tilde \cB + \tilde \cB\}.
\end{align}
Denote $\tilde \cB$-matrices as matrices with entries in set $\tilde \cB$. 
For $\tilde \cB$-matrices $\tilde \bA =[a_{i,j}]$ and $\tilde \bC = [c_{i,j}]$, we define $\tilde \cB$-matrix addition as follows:
\begin{align} \label{additionalgebra}
\tilde \bA + \tilde \bC = [a_{i,j} + c_{i,j}] \in \tilde \cB.
\end{align}
We define $\tilde \cB$-matrix multiplication as follows:
\begin{align} \label{productalgebra}
\tilde \bA \tilde \bC = [ \sum_{l} a_{i,l} \circ  c_{l,j}] \in \tilde \cB.
\end{align}
Due to the  non-linearity of elements in the matrices, the familiar matrix multiplication rules on scalar numbers (associativity, commutativity, and distribution) must be carefully applied to $\tilde \cB$-matrices. For $\tilde \cB$-matrix multiplications, we must strictly follow the canonical order of operations from right to left. For example, the order by which to evaluate $\bA\bB\bC$ is $(\bA(\bB\bC))$. Counterexamples can be provided to assert the following rules of $\tilde \cB$-matrix algebra: 
\begin{itemize}
\item $\bA\bB \neq \bB\bA$ (non-commutativity).
\item $\bA(\bB\bC) \neq (\bA\bB)\bC$ (non-associativity).
\item $\bA(\bB + \bC) \neq (\bA\bB + \bA\bC)$ (non-left-distributivity).
\item $c\bA \neq \bA c$ where $c$ is a scalar.
\end{itemize}
Nevertheless, the right-distributivity rule holds:
\begin{itemize}
\item $(\bA + \bB) \bC = \bA\bC + \bB\bC$.
\end{itemize}

\subsection{DAG-DNNs with addition-nodes}

\begin{figure}[h]
\centering
{
\subfigure[]
{\includegraphics[width=0.27\columnwidth]{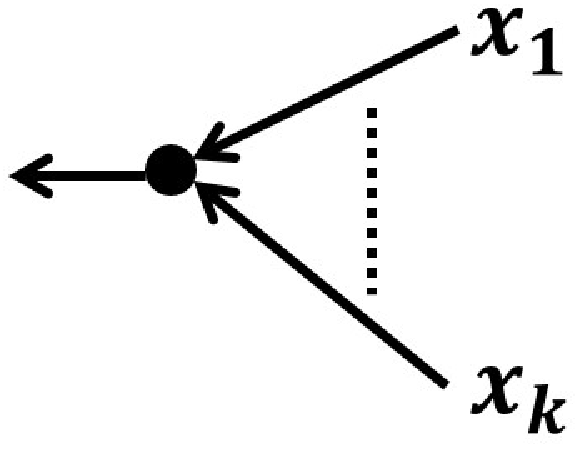}}
\subfigure[]{
\includegraphics[width=0.33\columnwidth]{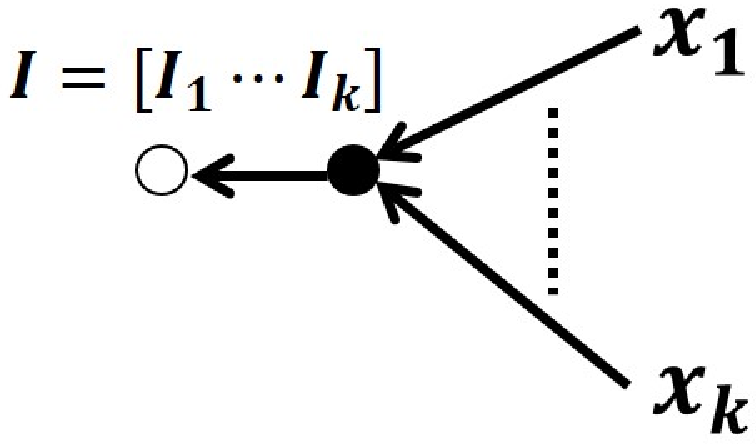}}
\subfigure[]{
\includegraphics[width=0.27\columnwidth]{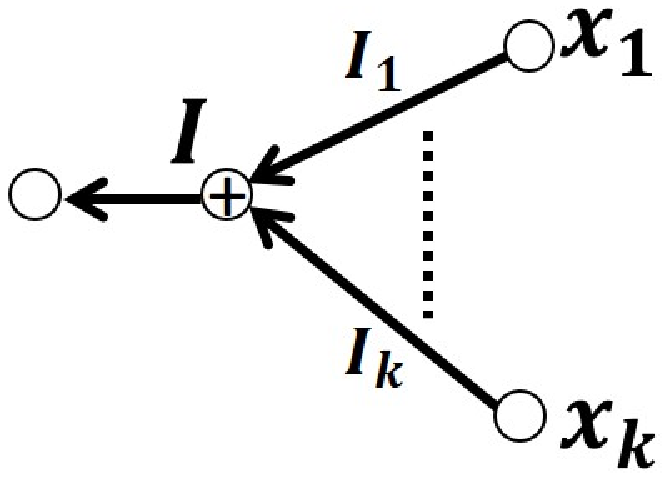}}}
\caption{In accordance with (\ref{addition}), concatenation operation (a) is replaced with addition operation (c); (b) the arc to the output of concatenation node (solid) is associated with identity matrix $\bI = [\bI_1, \cdots, \bI_k]$; (c) the concatenation node and its inputs are replaced with the addition-node ($\oplus$) with inputs $\{\bI_i \bx_i\}$ (apart from the addition-node, there are $k+1$ additional nodes in (c)). 
} \label{concatenationtransfer}
\end{figure}

Clearly, the atomic operations of series-connection and duplication can be expressed in the form of $\tilde \cB$-matrix addition and multiplication; however, using $\tilde \cB$-matrix addition and multiplication to express the atomic operation of concatenation is more involved. This requires the use of addition-nodes, which perform an  addition operation by taking $k$ vectors as inputs and outputting their sum:
\begin{align}
\oplus: \{\by_1, \cdots, \by_k\}\rightarrow \sum_{i=1}^k \by_i.
\end{align}
The operation of concatenation can then be replaced with the operation of addition using the algebra defined for $\tilde \cB$-matrices, as follows: 
\begin{align} \label{addition}
\begin{bmatrix}
\bx_1 \\
\vdots \\
\bx_k
\end{bmatrix}
= \bI 
\begin{bmatrix}
\bx_1 \\
\vdots \\
\bx_k
\end{bmatrix} 
= [\bI_1, \cdots, \bI_k] \begin{bmatrix}
\bx_1 \\
\vdots \\
\bx_k
\end{bmatrix}  = 
\sum_{i=1}^k \bI_i \bx_i = \sum_{i=1}^k \bI_i \circ \bx_i,
\end{align}
where $\bI_i$ is the restriction of identify matrix $\bI$ over columns corresponding to block $\bx_i$. The last equation relies on the fact that matrix multiplication is equivalent to matrix composition. Equation (\ref{addition}) is illustrated in Fig. \ref{concatenationtransfer}.

Throughout the rest, $\mathcal M^{\oplus}$ denotes the DAG-DNN $\mathcal M$ with addition-nodes, which is obtained by replacing all concatenation nodes in $\mathcal M$ with addition-nodes. 

\begin{figure}[h]
\centering
{
\subfigure[]{\includegraphics[width=0.4\columnwidth]{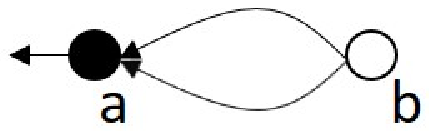}}}
{
\subfigure[]{
\includegraphics[width=0.4\columnwidth]{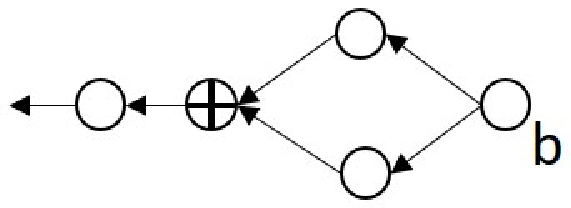}}}
\caption{(a) Node $a$ as a concatenation node; (b) Equivalent network corresponding to (a). } \label{singlelink}
\end{figure}

\begin{citedlem} \label{onearc}
Let $\mathcal M^\oplus$ be a DAG-DNN with addition-nodes. \\
(i) For any two nodes, there is no more than one arc from one node to the other node. \\ 
(ii) With the exception of input node $I$, precisely one arc is incident to any non-addition-node. 
\end{citedlem}
\proof
(i) 
As shown in Fig. \ref{singlelink}, all arcs incidental to node $a$ are replaced by an addition-node and the arcs incidental to the addition-node according to (\ref{addition}) are from different nodes.

%
%
(ii) The input node has no incident arc. 
In accordance with atomic operations $O1-O3$, only the concatenation nodes in $\mathcal M$ can have more than one incident arc and they have been replaced by addition-nodes in $\mathcal M^\oplus$. 
\qed

\begin{figure}[h]
\centering
{
\subfigure[]{\includegraphics[width=0.45\columnwidth]{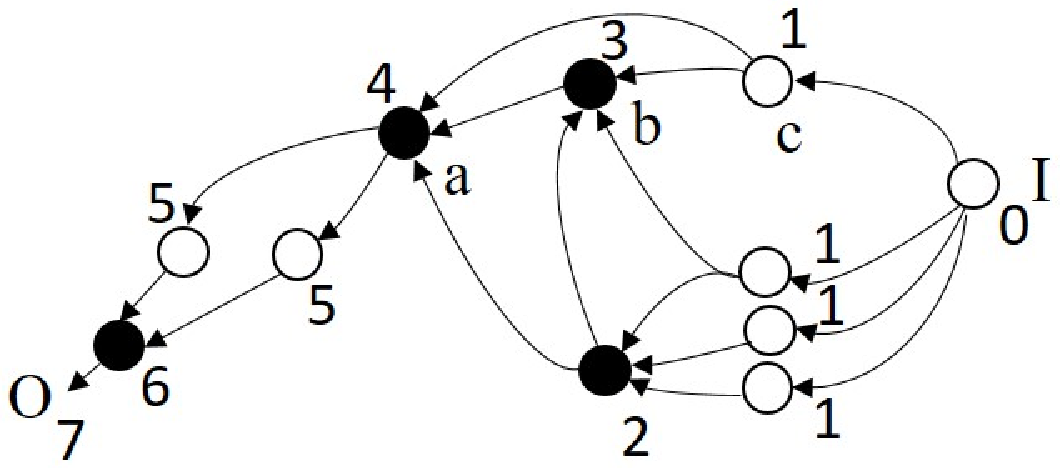}}}
{
\subfigure[]{
\includegraphics[width=0.45\columnwidth]{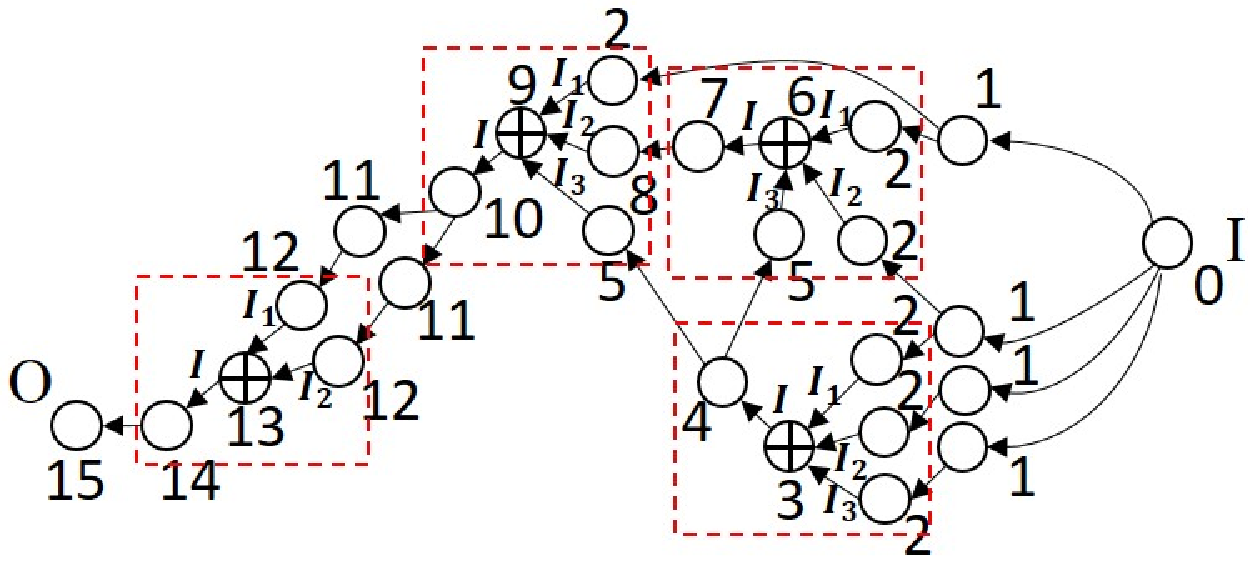}}}
{
\subfigure[]{
\includegraphics[width=0.45\columnwidth]{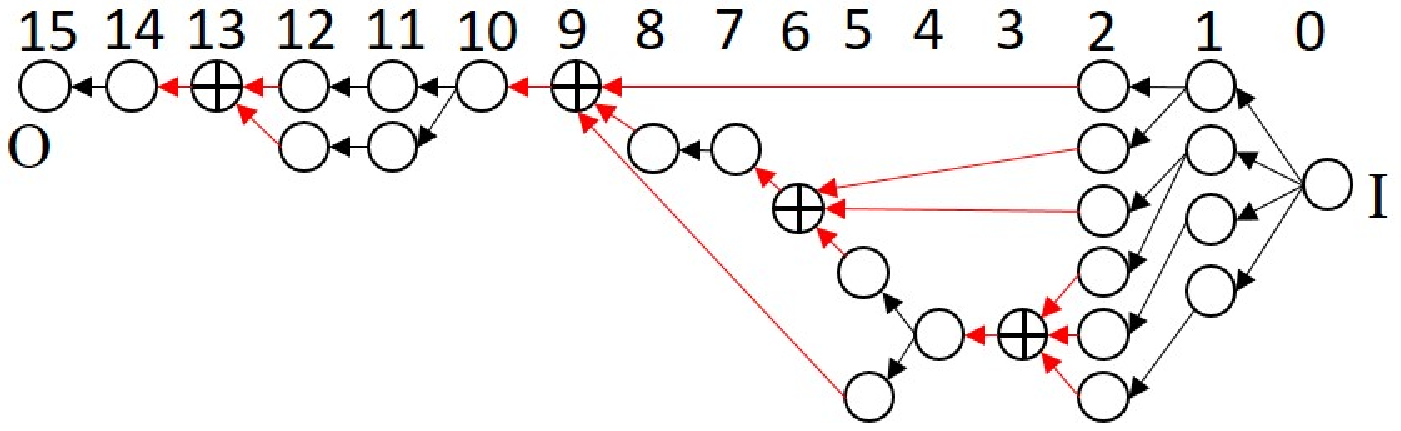}}}
{\subfigure[]{
\includegraphics[width=0.45\columnwidth]{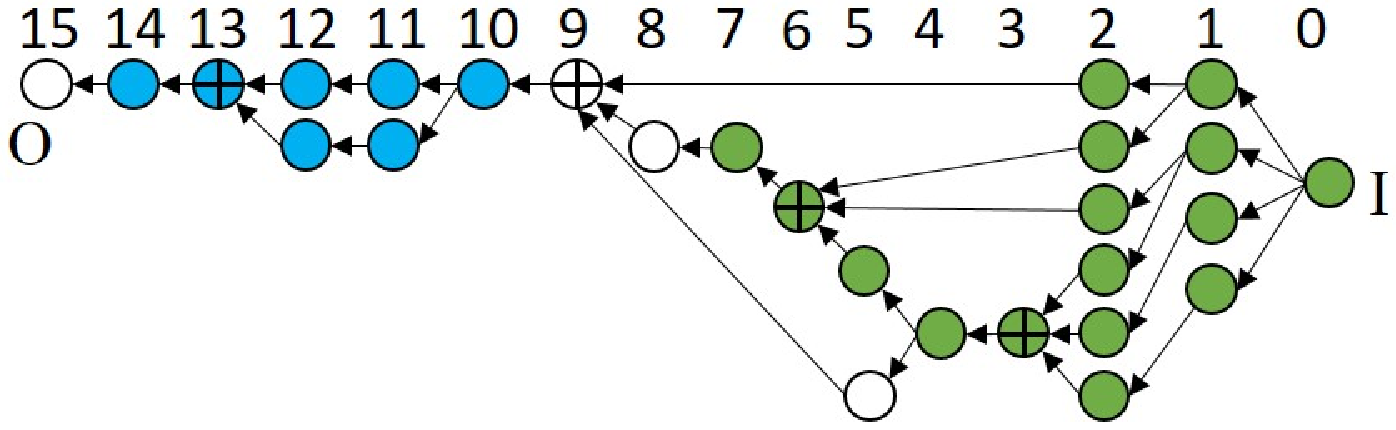}}}
\caption{(a) DAG-DNN $\mathcal M$ and (b) DAG-DNN $\mathcal M^{\oplus}$, where the dashed boxes indicate the replacement of concatenation nodes; (c) Level-graph of (b), where nodes are ordered according to their levels (0-15) (inputs and outputs to addition-nodes shown in red); (d) The green nodes form the complete sub-graph of the node at level $7$, whereas blue nodes form the complete sub-graph from the node at level $10$ to the node at level $14$ (note the sub-graph from the node at level $8$ to the node at level $9$ is not a complete sub-graph, because the addition-node at level $9$ is not complete within the sub-graph). The number next to a node in each sub-figure indicates its level. 
} \label{levelgraph}
\end{figure}

Fig. \ref{levelgraph}(a) and (b) respectively present $\mathcal M$ and $\mathcal M^{\oplus}$. 
It is possible to order nodes in a DAG-DNN based on their levels, defined as the number of arcs in the longest path between input node $I$ and the target node. Note that level here refers to a continuous integer between $0$ (the level of input node $I$) and $L$ (the level of output node $O$). The level-graph is a representation of DAG-DNN with addition-nodes where nodes are arranged in increasing order based on their respective levels. The level-graph for Fig. \ref{levelgraph}(b) is shown in Fig. \ref{levelgraph}(c) with the notation relevant to level-graphs listed below.

\subsection{Matrix compositions of level-graphs}

A DAG-DNN with addition-nodes can be ordered according to the levels of nodes into a level-graph from which it is possible to derive a representation using matrix composition.

\begin{citeddef} \label{def-level-graph}
(i) $\text{Level}(a)$ denotes the level of node $a$ (i.e., the number of arcs in the longest path between input node $I$ and node $a$). \\
(ii) $N_n$ denotes the nodes at level $n$, $N_0 = N_{\leq 0} = I$ (the input node), and $N_{\leq n}  = [N_{\leq n-1}^\top \; N_{n}^\top]^\top$ denotes the collection of nodes up to level $n$, where $n \geq 1$. \\
(iii) $m_l$ denotes the number of nodes at level $l$, and $m_{\leq n}$ denotes the  number of nodes in $N_{\leq n}$ (hence, $m_{\leq n} = \sum_{l=0}^n m_l$). \\
(iv) Arc $a \rightarrow b$ is a jump if and only if $\text{level}(b) - \text{level}(a) \geq 2$. 
\end{citeddef}
Obviously, for a DNN of $L+1$ levels, input node $I$ is at level $0$ and output node $O$ is at level $L$; i.e., $m_0 = 1$ and $m_{L} = 1$.

Deducing that the end node of a jump in $\mathcal M^\oplus$ must be an addition node is straightforward. If jump $a_0 \rightarrow a_n$ with $\text{level}(a_n) - \text{level}(a_0) = n$, then we can replace the jump with the following chain $a_0 \rightarrow a_1 \rightarrow \cdots \rightarrow a_{n-1} \rightarrow a_n$. This allows us to attach the function on the jump to arc $a_{0} \rightarrow a_1$, and attach an identify function to any other arc in the chain. Node $a_i$ is added to level $\text{level}(a_0) + i$ with $i=1, \cdots, n-1$. Replacing all jumps in $\mathcal M^\oplus$ yields a level-graph comprising arcs that involves nodes only at adjacent levels, which means that the outputs of nodes at one level are inputs to the nodes at the next level. Let $\bA_{l+1,l} \in \bB^{m_{l+1} \times  m_l}$ be the matrix of functions to arcs from nodes at level $l$ to nodes at level $l+1$, where $\bA_{l+1,l}(b,a)$ is the function attached to arc $a \rightarrow b$. Let $\mathcal M^{\oplus}$ be the DAG-DNN in which all concatenation nodes were replaced by addition-nodes and all jumps were replaced by chains. The network can then be delineated according to level, as follows:
\begin{align} \label{representationA}
\mathcal M^{\oplus} = \bA_{L, L-1} \bA_{L-1, L-2} \cdots \bA_{1, 0}.
\end{align}
The multiplication and addition respectively obey the rules of Equations (\ref{additionalgebra}) and (\ref{productalgebra}) for a network evaluation at input $\bx$, as follows:
\begin{align} \label{univmatrix}
\mathcal M^{\oplus}(\bx)  =  \bA_{L, L-1} \bA_{L-1, L-2} \cdots \bA_{2, 1}\bA_{1, 0}(\bx). 
\end{align}
Network $M_L \circ \varrho_{L-1} \circ\cdots\circ\varrho_1\circ M_1(\bx)$ has corresponding $\bA_{l+1, l} = \varrho \circ M_{l+1} \in \bB$ for $l=0, \cdots L-2$ and $\bA_{L, L-1} = M_{L} \in \bB$.

\begin{figure}[th]
\begin{center}
  \subfigure[]{\includegraphics[width=0.3\textwidth]{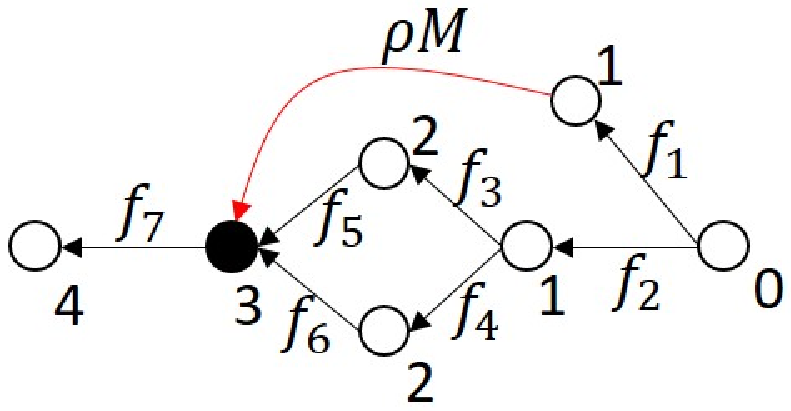}} 
   \subfigure[]{\includegraphics[width=0.4\textwidth]{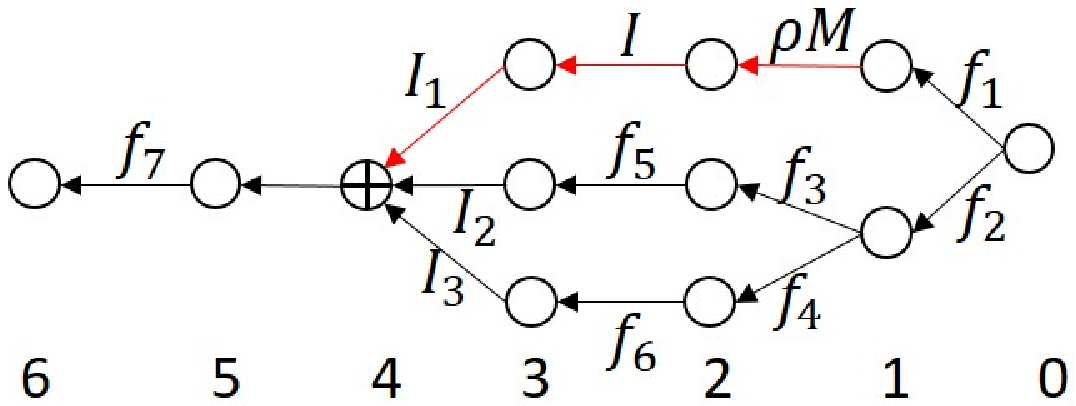}} 
  \end{center} \caption{
  (a) Network of one concatenation node (solid node) and a jump (arc in red); (b) The concatenation node was replaced with an addition-node and the jump was replaced by a chain (arcs in red). The numbers next to nodes indicate their corresponding levels.
  } \label{jump}
\end{figure}

Fig. \ref{jump}(b) presents the level-graph of Fig. \ref{jump}(a) after replacing the concatenation node and jump. The network in Fig. \ref{jump}(b) can be described as follows: 
\begin{align*}
[f_7] [\bI] \begin{bmatrix} \bI_1 & \bI_2 & \bI_3 \end{bmatrix} 
\begin{bmatrix} \bI & 0 & 0 \\ 0 & f_5 & 0 \\ 0 & 0 & f_6 \end{bmatrix}
\begin{bmatrix} \rho M & 0 \\ 0 & f_3 \\ 0 & f_4 \end{bmatrix} 
 \begin{bmatrix} f_1 \\ f_2 \end{bmatrix}
\end{align*}
where $\bA_{1, 0} =  \begin{bmatrix} f_1 \\ f_2 \end{bmatrix}$, $\bA_{2,1} = \begin{bmatrix} \rho M & 0 \\ 0 & f_3 \\ 0 & f_4 \end{bmatrix}$.  Using definition 
$\bA_{1,0} (\bx) = \begin{bmatrix} f_1(\bx) \\ f_2(\bx) \end{bmatrix}$,
the network evaluation at $\bx$ can be expressed as follows:
\begin{align*}
[f_7] [\bI] \begin{bmatrix} \bI_1 & \bI_2 & \bI_3 \end{bmatrix} 
\begin{bmatrix} \bI & 0 & 0 \\ 0 & f_5 & 0 \\ 0 & 0 & f_6 \end{bmatrix}
\begin{bmatrix} \rho M & 0 \\ 0 & f_3 \\ 0 & f_4 \end{bmatrix} 
 \begin{bmatrix} f_1(\bx) \\ f_2(\bx) \end{bmatrix}.
\end{align*}

Representation (\ref{representationA}), which is universal to all DAG-DNNs, depicts the function of $\mathcal M^\oplus$ in a useful and compact form; however, functions can be associated to sub-graphs of $\mathcal M^\oplus$ and they are obscure to the representation. We present in the next section an extension of (\ref{representationA}), which elucidates all functions on the sub-graphs of a DAG-DNN.

%

\section{Functions on sub-graphs of DAG-DNNs} \label{sec:representation}

\subsection{Compositions of lower-triangle matrices}


Denote $f_{\mathcal M^{\oplus}}[i,j]$ as the function restricted on sub-graph with input at node $j$ and output at node $i$. The input domain of the function is the output domain at node $j$. Here, restriction the domain on a sub-graph means that nodes not in the sub-graph are not considered, and inputs to an addition-node from the nodes not in the sub-graph are set to zero.
The transitivity of a connected DAG-DNN facilitates the definition of functions in sub-graphs via induction on the levels of nodes in level-graphs of $\mathcal M^\oplus$.

Initialization step: We let 
\begin{align} \label{funbasic0}
f_{\mathcal M^{\oplus}}[a,a] = \bI
\end{align}
where $\bI$ refers to an identify function indicating the reachability of any node $a$ to itself. 

Basic step (nodes at level $1$): Suppose that $a$ is at level $1$. In accordance with Lemma \ref{onearc}, node $a$ is not an addition-node, which means that $a \leftarrow I$ is a sub-graph of nodes $a$ and $I$, such that 
\begin{align}\label{funbasic}
f_{\mathcal M^{\oplus}}[a,I] \in \cB =
\text{arc-function}[a,I] \text{ if $a \leftarrow I$.} 
\end{align}

Induction step (nodes at levels $\geq 2$): 
Suppose that we obtain functions for sub-graphs using nodes up to level $k$, and further suppose that node $a$ is at level $k+1$.  In view of Lemma \ref{onearc}, and recalling that $\circ$ denotes the composition, 
we can determine via induction that the sub-graph between node $c$ and node $a$ is a union of sub-graphs between node $c$ and node $b$ with $a \leftarrow b$ and then between node $b$ and node $a$. If node $a$ is unreachable from node $c$ (i.e., the sub-graph is an empty graph), then we let $f_{\mathcal M^{\oplus}}[a, c] = \b0$. The following is applicable to any node $c$ at level below $k+1$:

Case 1: Node $a$ is not an addition-node (inlet of $a$ is $1$, Lemma \ref{onearc}):
We obtain the following:
\begin{align}\label{fundirect0}
f_{\mathcal M^{\oplus}}[a,c] \in \tilde \cB = 
\begin{cases}
\text{arc-function}[a,b] \circ f_{\mathcal M^{\oplus}}[b,c], \text{ if $a \leftarrow b$ and $b$ is reachable from $c$;} \\
\b0 \text{ if $a$ is unreachable from $c$.}
\end{cases}
\end{align}

Case 2: Node $a$ is an addition-node of $k$-inlets ($k \geq 2$):
We obtain the following:

\begin{align} \label{fundirect}
f_{\mathcal M^{\oplus}}[a,c] \in \tilde \cB =
\begin{cases}
  \sum_{\{b|a \leftarrow b\}} \text{arc-function}[a, b]\circ f_{\mathcal M^{\oplus}}[b,c], \text{ if $b$ is reachable from $c$;} \\
  \text{arc-function}[a, c], \text{ if $a \leftarrow c$};\\
   \b0 \text{ if $a$ is unreachable from $c$.}
 \end{cases}
\end{align}

We can form an all-pair function matrix of nodes with entries defined using Eqs. (\ref{funbasic0})-(\ref{fundirect}), such that sub-graphs of $\mathcal M^\oplus$ have corresponding functions in the matrix. The row and column indices of the matrix are consistent with the node levels in level-graphs.
We arrange nodes in increasing order based on their levels, such that the first $m_{\leq n}$ rows and columns of the matrix are associated with nodes in $N_{\leq n}$. 
We shall demonstrate that the matrix of all-pair functions can be factorized (decomposed) as multiplications of lower-triangle matrices, each of which records the transition to the next level with arc-function entries defined in (\ref{funbasic}). The decomposition can be deemed unique up to simultaneous permutations of rows and corresponding columns in the matrix. This is equivalent to re-numbering the nodes in a graph. 

Denote $\mathcal M_{\leq n}^\oplus$ as the sub-graph of $\mathcal M^\oplus$, wherein the nodes are the same as $\mathcal M^\oplus$, but arcs incidental to any nodes at levels higher than $n$ are removed. Further denote $\bar{\mathcal M}^\oplus_{\leq n}$ as the sub-graph of $\mathcal M_{\leq n}^\oplus$ by restricting the nodes in $\mathcal M_{\leq n}^\oplus$ at levels up to $n$ (i.e., removing all nodes at levels above $n$ in $\mathcal M^\oplus_{\leq n}$).

\begin{citeddef} \label{allpairsdef}
(i) All-pair function matrix $\bC=\bC_{\leq L}$ for $\mathcal M^{\oplus}$ is a $m\times m$ ($m=m_{\leq L}$) square $\tilde \cB$-matrix with the following entries:
\begin{align}
\bC_{\leq L}[i,j] = f_{\mathcal M^{\oplus}}[i,j].
\end{align}
(ii) $\bC_{0}$ is an identity matrix of size $m \times m$. \\
(iii) $\bC_{\leq n} \in \tilde \cB^{m \times m}$ denotes the all-pair function matrix for $\mathcal M_{\leq n}^\oplus$, wherein the block $\bC_{\leq n}[1:m_{\leq n}, 1:m_{\leq n}] = \bC_{\leq L}[1:m_{\leq n}, 1:m_{\leq n}]$. The entries of $\bC_{\leq n}$ are zeros on the off-diagonal entries involving nodes at levels above $n$:
\begin{align} \label{functionlevel}
\bC_{\leq n}[i,j] \in 
\begin{cases} \bC_{\leq L}[i,j] \text{ if $n \geq$ level($i$), level($j$)}, \\
\bI \text{ if $i = j$ for level($i$) $> n$,} \\
\b0 \text{ otherwise.}
\end{cases}
\end{align}
(iv) Denote $\bar \bC_{\leq n} \in \tilde \cB^{m_{\leq n} \times m_{\leq n}}= \bC_{\leq n}[1: m_{\leq n}, 1: m_{\leq n}]$ as the all-pair function matrix for $\bar{\mathcal M}^\oplus_{\leq n}$. This means that $\bC_{\leq n}$ can be expressed as block matrix, as follows:
\begin{align} \label{transitivitymatrix}
\bC_{\leq n}= \begin{bmatrix} \bar \bC_{\leq n} & \b0 \\
\b0 & \bI
\end{bmatrix},
\end{align}
wherein $\b0$ denotes zero matrices of appropriate sizes. Note that $\bar \bC_{0} = \bar \bC_{\leq 0}$ and
$\bC_0 = \bC_{\leq 0}$. \\
(v) We denote $\bE_{n, \leq n-1} \in \cB^{m_n \times m_{\leq n-1}}$ as the matrix of functions on arcs incidental to nodes at level $n$ from any nodes at level $\leq n-1$: 
\begin{align} \label{fdefinition}
\bE_{n, \leq n-1}[i,j] = 
\begin{cases}
\text{arc-function}[i,j]  \text{ if $i \leftarrow j$;}  \\
\b0 \text{ if $i$ is unreachable from $j$;}
\end{cases}
\end{align}
where node $j$ is at a level $\leq n-1$ and node $i$ is at level $n$. 
\end{citeddef}

The introduction of $\bE_{n, \leq n-1}$ allows us to express $\bar \bC_{\leq n}$ in (\ref{transitivitymatrix}) in block matrix form as follows:
\begin{align} \label{functiontransitivity}
\bar \bC_{\leq n} = \begin{bmatrix} \bar \bC_{\leq n-1} & \b0 \\
\bE_{n, \leq n-1} \bar \bC_{\leq n-1}& \bI \end{bmatrix}.
\end{align}
The top-left block corresponds to the all-pair function matrix of $\bar{\mathcal M}^\oplus_{\leq n-1}$ (i.e., the sub-graphs of nodes up to level $n-1$). The bottom-left block represents the functions on sub-graphs between nodes at level $n$ and at level $< n$. Matrix multiplication is implemented on elements in $\cB$ and $\tilde \cB$ in accordance with the rules in  (\ref{additionalgebra}) and  (\ref{productalgebra}).

We introduce the (square) lower-triangle matrix $\bB_{n, n-1} \in \cB^{m \times m}$ to record the transition from nodes at level $\leq n-1$ to level $n$ in $\mathcal M^\oplus$ with   
\begin{align} \label{tworelation1}
\bB_{n, n-1} =   \begin{bmatrix} \bI_{m_{\leq n-1}} &  \b0 & \b0\\
 \bE_{n, \leq n-1} & \bI_{m_{n}} & \b0 \\
 \b0 & \b0 & \bI \end{bmatrix}.
\end{align}
Below, we show that an all-pair function matrix is invertible and can be decomposed as multiplications of lower-triangle matrices. The following theorem is universally applicable to any DAG-DNN.

\begin{citedthm} (Lower-triangle factorization)  \label{universal}
Given an $L+1$-level $\mathcal M^\oplus$ (i.e., level $0$-$L$), the corresponding all-pair function matrix $\bC_{\leq L}$ can be expressed as multiplications of lower-triangle matrices $\{\bB_{n+1, n} \in \cB^{m \times m}\}_{n=0}^{L-1}$ with the following form:
\begin{align} \label{tworelation}
\bC_{\leq L}  = \bB_{L,L-1} \bC_{\leq L-1} = \bB_{L,L-1}\cdots \bB_{1,0} \bC_0.
\end{align}
The square matrix $\bB_{n+1, n}$ is given in (\ref{tworelation1}) and $\bC_0 = \bI_m$ (Definition \ref{allpairsdef}).
Moreover, $\bB_{n+1, n}$ is invertible using the following invertible matrix:
\begin{align} \label{inverselifting}
\bB^{-1}_{n+1, n} =   \begin{bmatrix} \bI_{m_{\leq n}} &  \b0 & \b0\\
 -\bE_{n+1, \leq n} & \bI_{m_{n+1}} & \b0 \\
 \b0 & \b0 & \bI \end{bmatrix}.
\end{align}
Hence, $\bC_{\leq L}$ is invertible with $\bC_{\leq L}^{-1} \bC_{\leq L} = \bC_{\leq L} \bC^{-1}_{\leq L} = \bI_m$, where 
\begin{align} \label{inverform}
\bC_{\leq L}^{-1}  = \bC^{-1}_{\leq L-1} \bB^{-1}_{L,L-1} =  \bB^{-1}_{1,0}\cdots \bB^{-1}_{L,L-1} .
\end{align}
\end{citedthm}
\proof
It suffices to show that
\begin{align} \label{twolevelgraphs}
\bC_{\leq n+1} = \bB_{n+1, n} \bC_{\leq n}.
\end{align}
This can be straightforwardly derived by showing that multiplication of $\bB_{n+1, n}$ with $\bC_{\leq n}$, expressed as a $3 \times 3$ block matrix:
\begin{align*}
\bC_{\leq n} = \begin{bmatrix}
\bar \bC_{\leq n} & \b0 & \b0 \\
\b0_{m_{n+1} \times m_{\leq n}} & \bI_{m_{n+1}} & \b0 \\
\b0 & \b0 & \bI
\end{bmatrix}
\end{align*}
yields
\begin{align} \label{transitiveto}
\bC_{\leq n+1} = 
\begin{bmatrix}
\bar \bC_{\leq n} & \b0 & \b0 \\
 \bE_{n+1, \leq n} \bar \bC_{\leq n}  & \bI_{m_{n+1}} & \b0 \\
  \b0 & \b0 & \bI
\end{bmatrix}.
\end{align}
$\bB_{n+1, n}^{-1}$ is the invertible matrix of $\bB_{n+1, n}$ because 
\begin{align*}
\begin{bmatrix}
\bI_{m_{\leq n}} & \b0 &\b0 \\
\b0_{m_{n+1}, m_{\leq n}} & \bI & \b0 \\
\b0 & \b0 & \bI 
\end{bmatrix} = \begin{bmatrix}
\bI_{m_{\leq n}} & \b0 & \b0 \\
-\bE_{n+1, \leq n} & \bI & \b0 \\
\b0 & \b0 & \bI 
\end{bmatrix} \begin{bmatrix}
\bI_{m_{\leq n}} &\b0 & \b0 \\
\bE_{n+1, \leq n} & \bI & \b0 \\
\b0 & \b0 & \bI 
\end{bmatrix} = \bB_{n+1, n}^{-1}\bB_{n+1, n}.
\end{align*}
Thus, $\bC_{\leq L}$ is an invertible matrix and (\ref{inverform}) is immediately obtained.

\qed

\begin{figure}[th]
\begin{center}
  \subfigure[$\mathcal M$]{\includegraphics[width=0.28\textwidth]{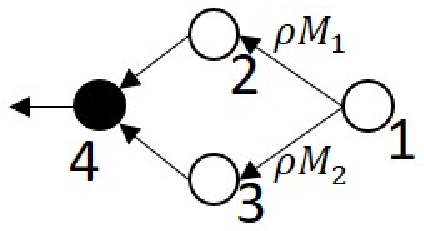}}
  \subfigure[$\mathcal M^\oplus$]{\includegraphics[width=0.28\textwidth]{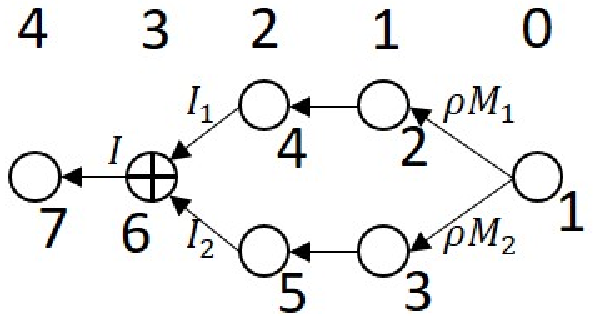}} 
  \end{center} \caption{
  (a) DAG-DNN comprising four nodes, among which node $4$ is the concatenation output node and arcs $2 \rightarrow 4$ and $3 \rightarrow 5$ are identity functions; (b) Equivalent $5$-level DAG-DNN comprising one addition-node and seven nodes with levels numbered along the top row. 
  } \label{stacking0}
\end{figure}

\begin{figure}[h]
\begin{center}
  \subfigure[$\bB_{1,0}$]{\includegraphics[width=0.28\textwidth]{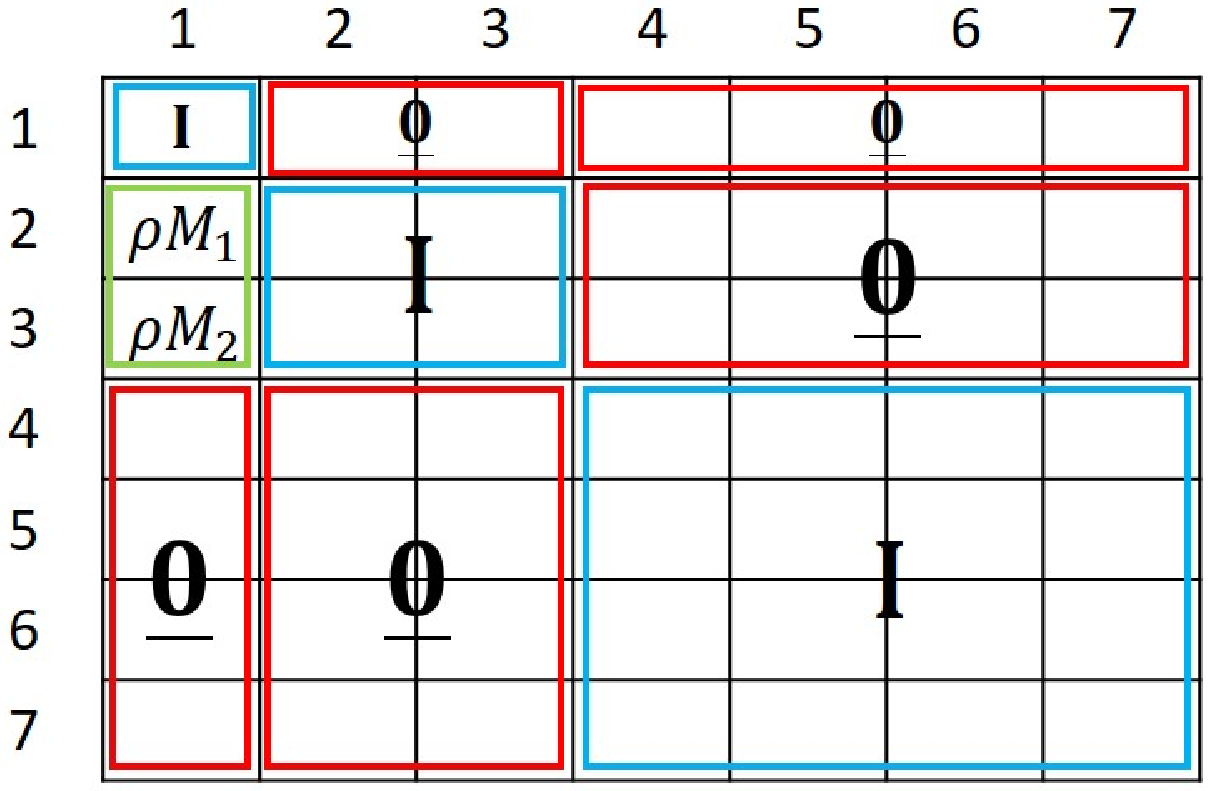}}
  \subfigure[$\bB_{2,1}$]{\includegraphics[width=0.28\textwidth]{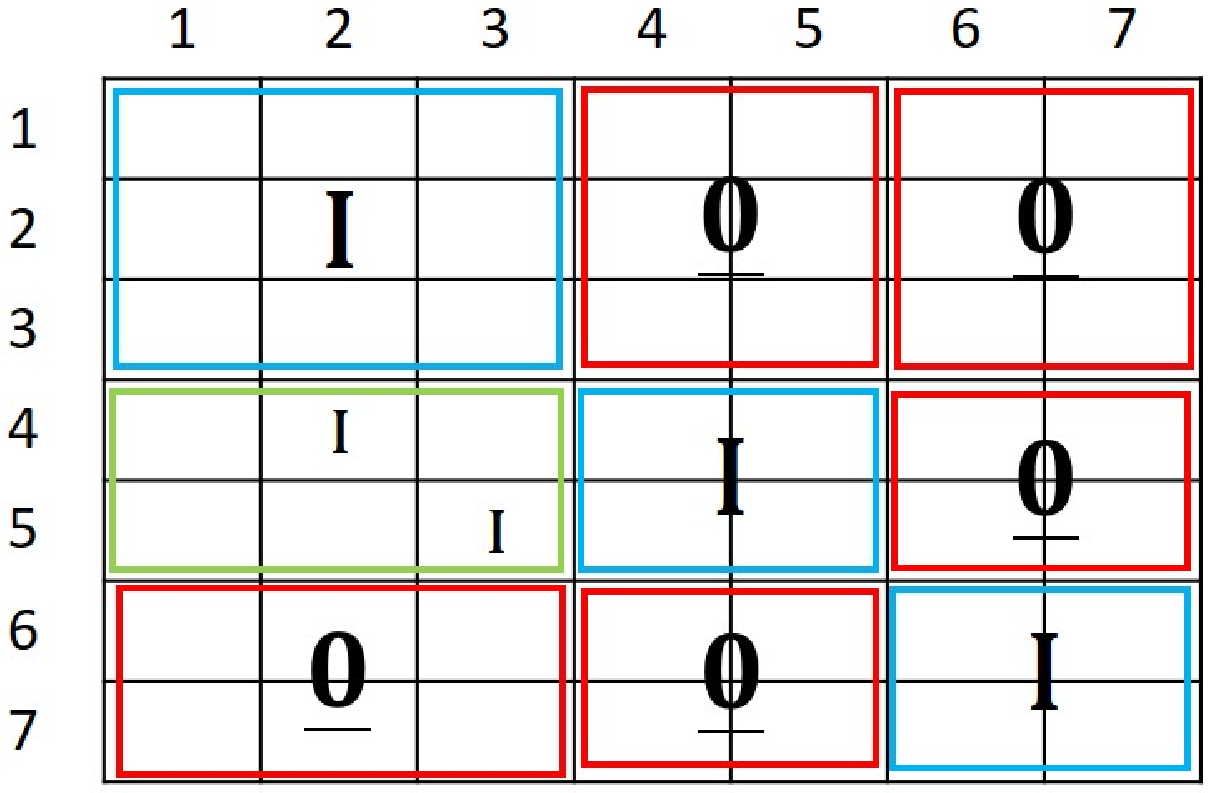}} 
  \subfigure[$\bB_{3,2}$]{\includegraphics[width=0.28\textwidth]{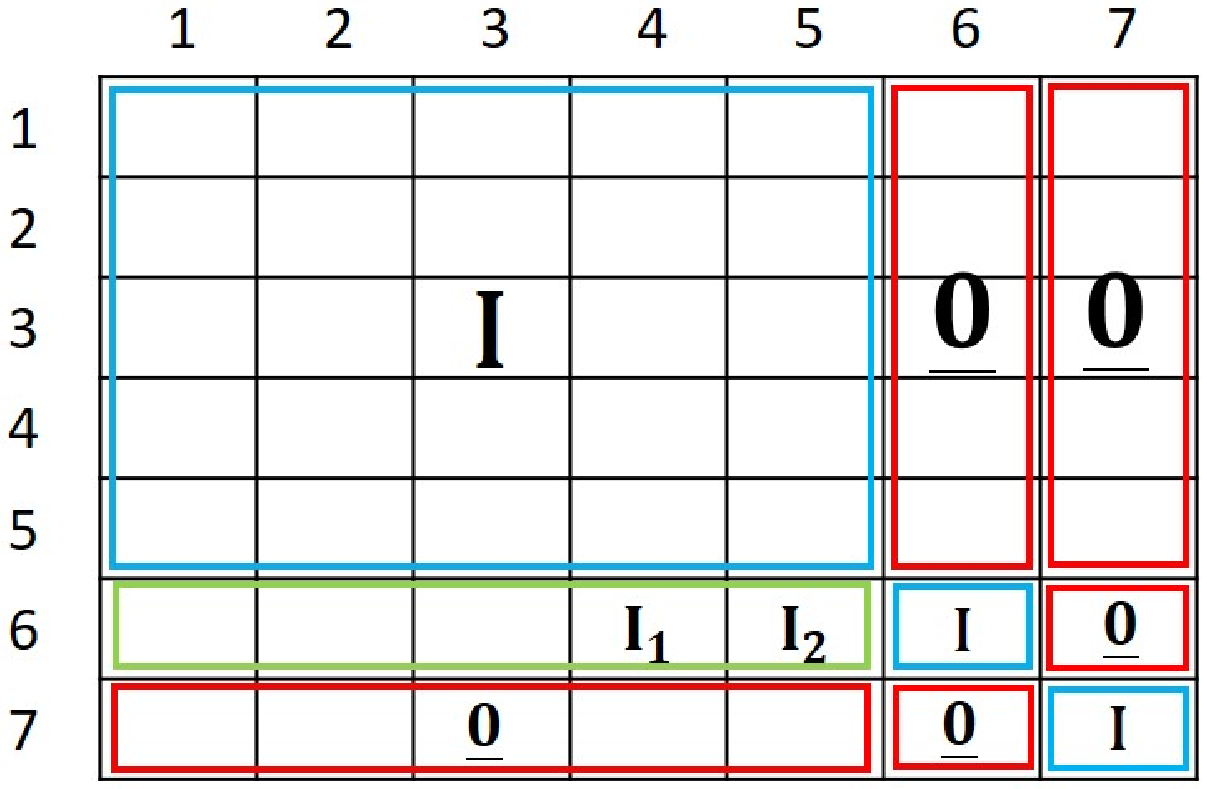}}
  \subfigure[$\bB_{4,3}$]{\includegraphics[width=0.28\textwidth]{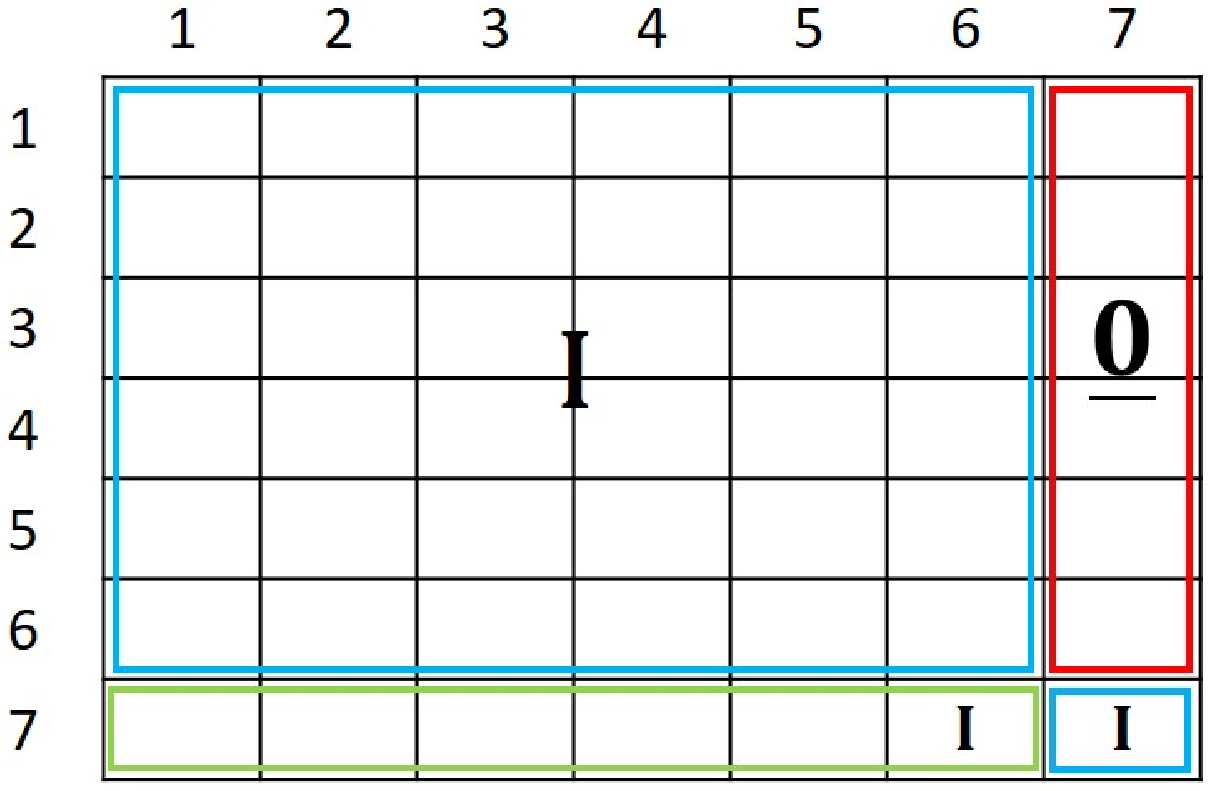}} 
    \subfigure[$\bC_{\leq 1}$]{\includegraphics[width=0.28\textwidth]{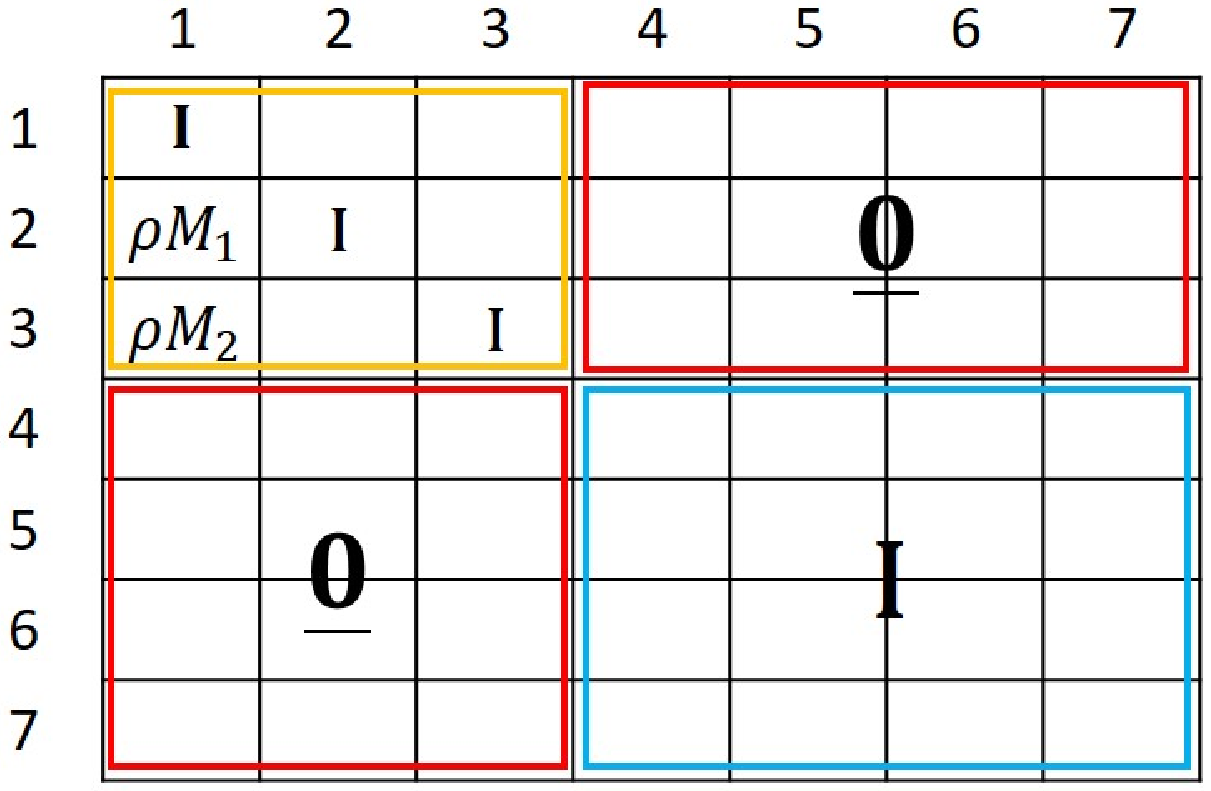}} 
  \subfigure[$\bC_{\leq 2}$]{\includegraphics[width=0.28\textwidth]{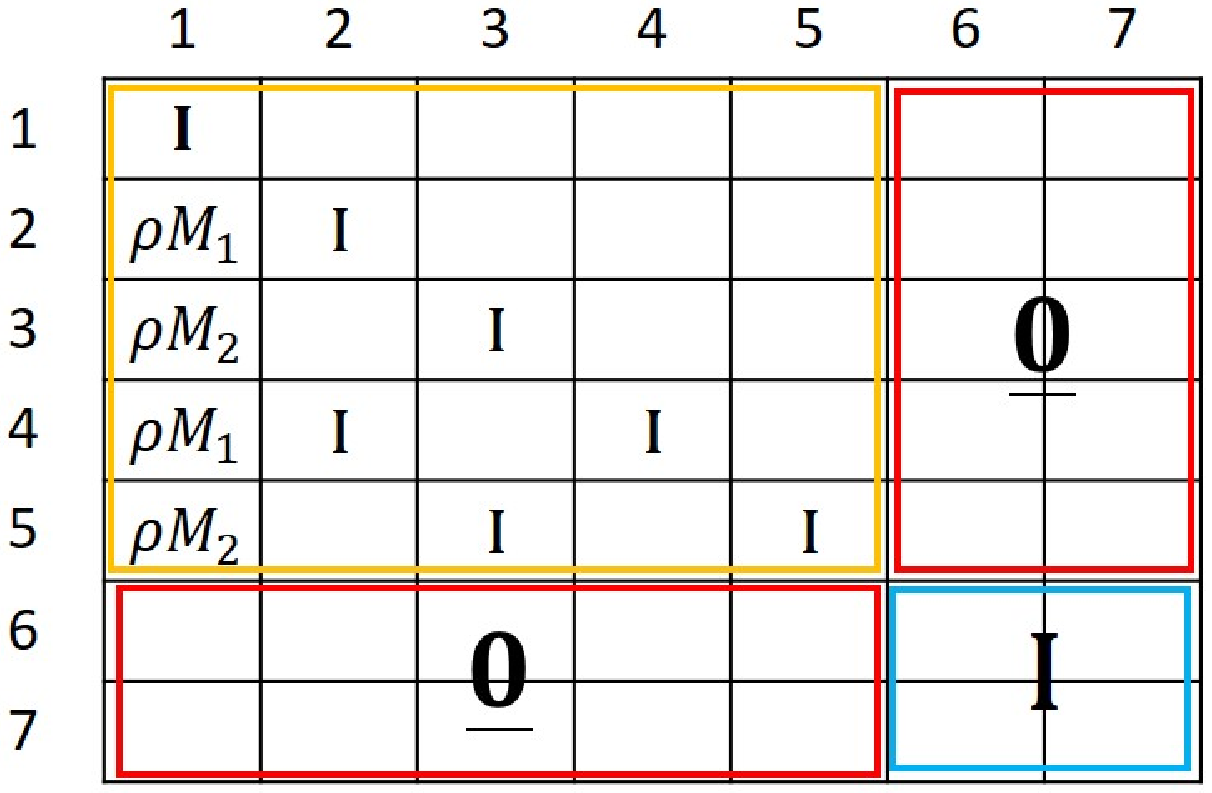}} \vspace{-0.1in}\\
  \subfigure[$\bC_{\leq 3}$]{\includegraphics[width=0.30\textwidth]{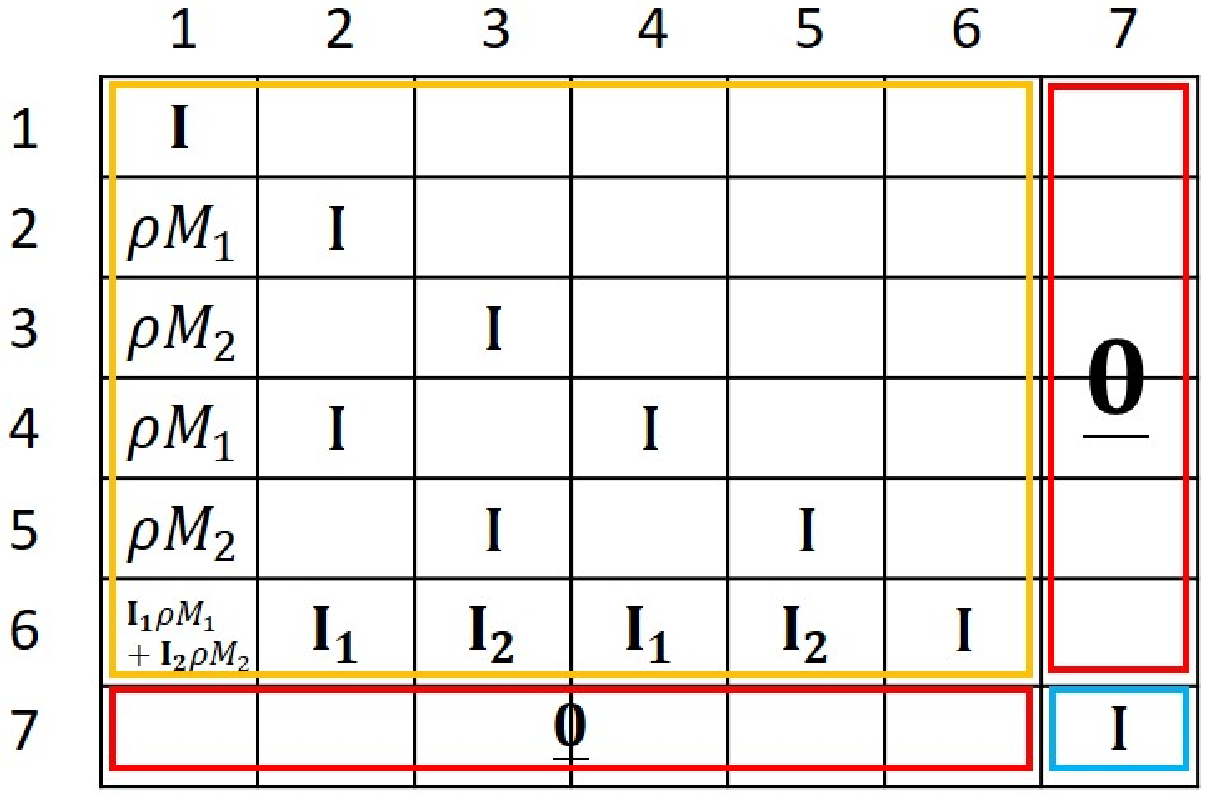}}
  \subfigure[$\bC_{\leq 4}$]{\includegraphics[width=0.30\textwidth]{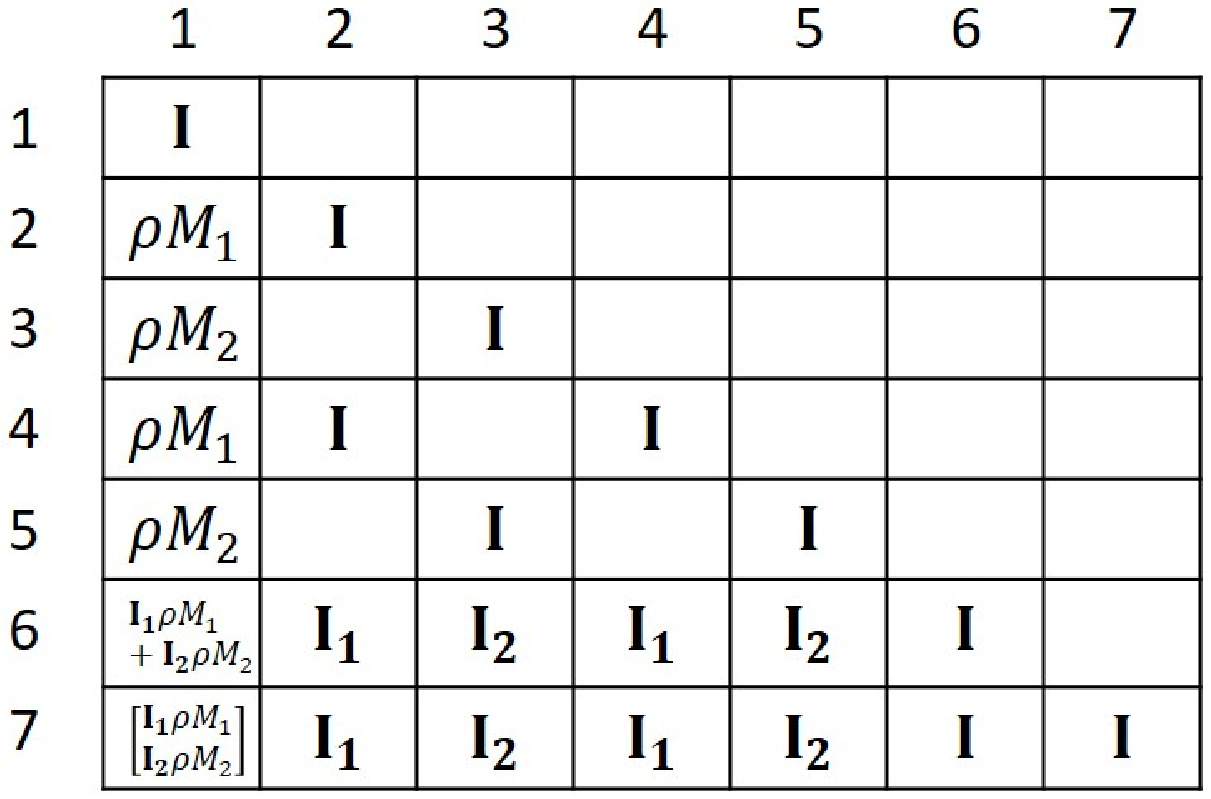}} \vspace{-0.1in} \\
  \subfigure[Complete $\bC_{\leq 3}$ ]{
\includegraphics[width=0.30\columnwidth]{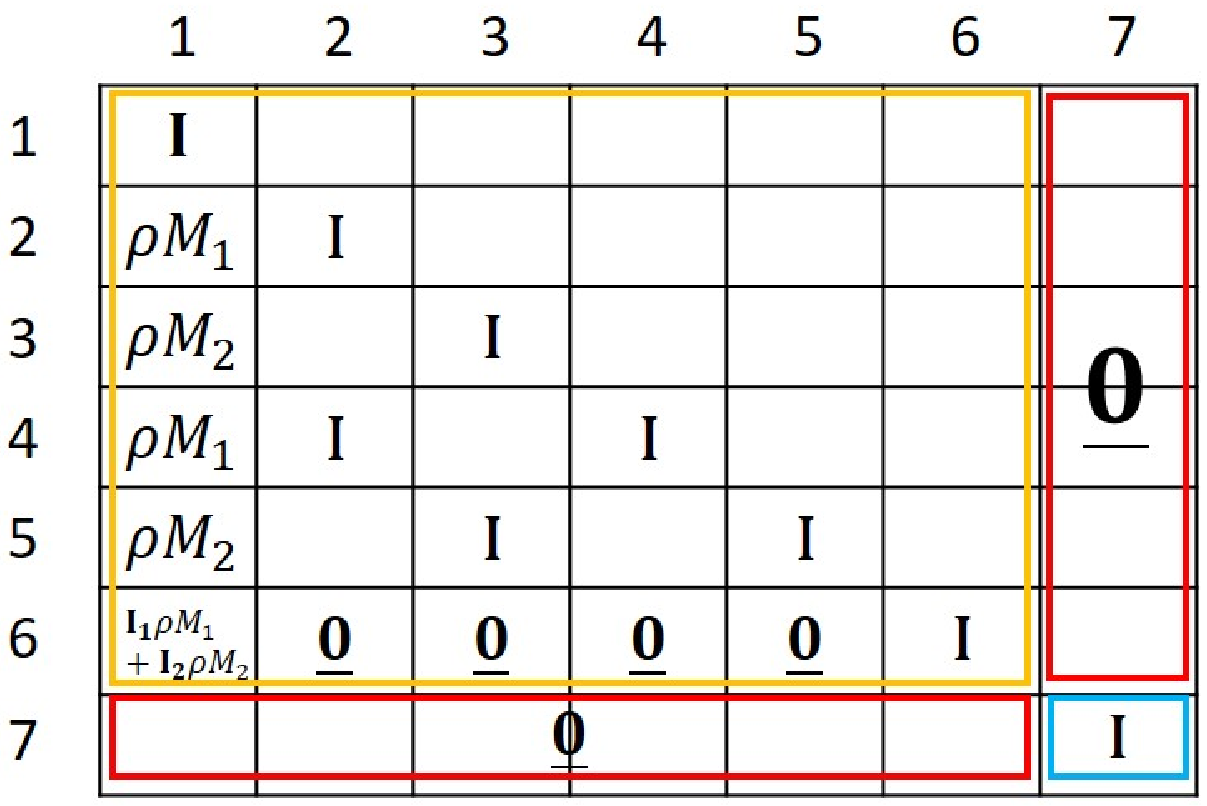}}
\subfigure[Complete $\bC_{\leq 4}$ ]{
\includegraphics[width=0.30\columnwidth]{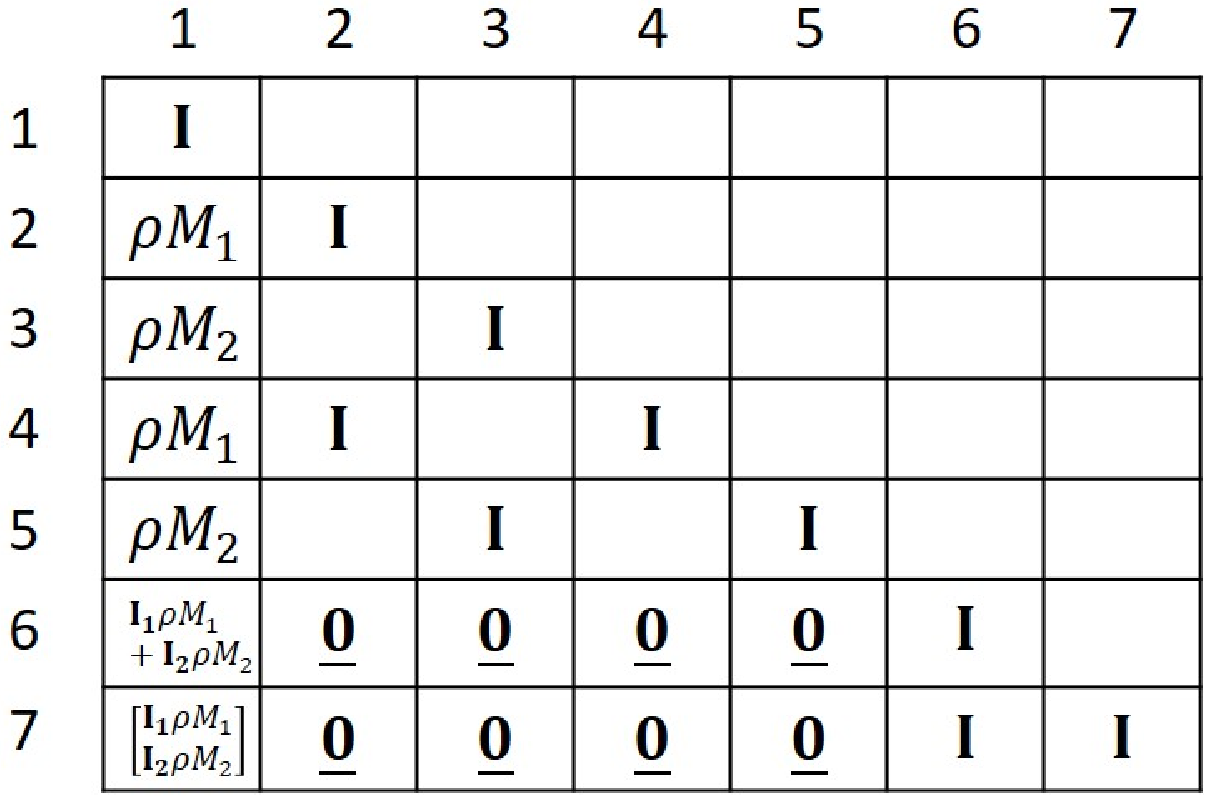}}
\end{center}\caption{
(a)-(d) $\bB_{1,0}-\bB_{4,3}$ for Fig. \ref{stacking0}(b); (e)-(h) $\bC_{\leq 1}-\bC_{\leq 4}$; (i) and (j) complete sub-graphs up to level $3$ and $4$, respectively. 
}
\label{stacking}
\end{figure}

Fig. \ref{stacking}(h) presents the all-pair functions of DAG-DNN $\mathcal M$ in Fig. \ref{stacking0}(a). To this end, $\mathcal M$ is first transformed into $\mathcal M^\oplus$ in Fig. \ref{stacking0}(b). We then derive Fig. \ref{stacking}(h) recursively (e.g., the all-pair function matrix $\bC_{\leq 4}$ is derived from $\bC_{\leq 3}$ and $\bB_{4, 3}$).
Figs. \ref{stacking}(a)-(d) present the respective $\bB_{n+1, n}$ for $n=0, 1, 2, 3$. Figs. \ref{stacking}(e)-(h) present $\bC_{\leq n}$ for $n=1, 2, 3, 4$.
Node ($6$) in Fig. \ref{stacking0}(b) is an addition-node. Entries $(6,1)$ in Figs. \ref{stacking} (g) and \ref{stacking}(h) are additions of functions deduced from paths $1\rightarrow 2\rightarrow 4 \rightarrow 6$ and $1\rightarrow 3\rightarrow 5 \rightarrow 6$. Entry $(7,1)$ in Fig. \ref{stacking}(h) is the output of $\mathcal M^\oplus$. Note that Figs. \ref{stacking}(a)-(g) are block matrices and their boundaries are displayed in colors for easy delineation.

\begin{figure}[th]
\centering
{
\subfigure[]{
\includegraphics[width=0.36\columnwidth]{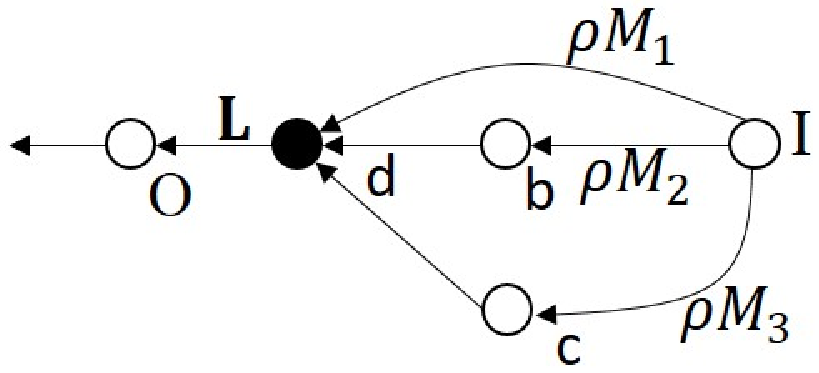}} 
\subfigure[]{
\includegraphics[width=0.36\columnwidth]{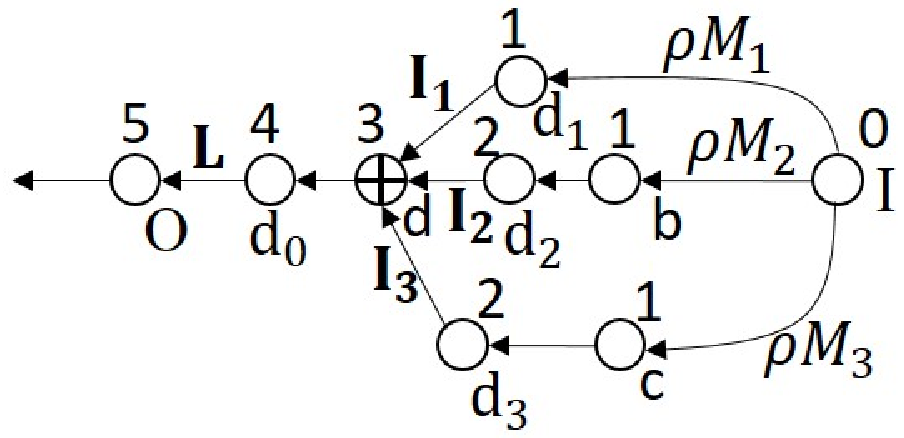}} \\
\subfigure[]{
\includegraphics[width=0.36\columnwidth]{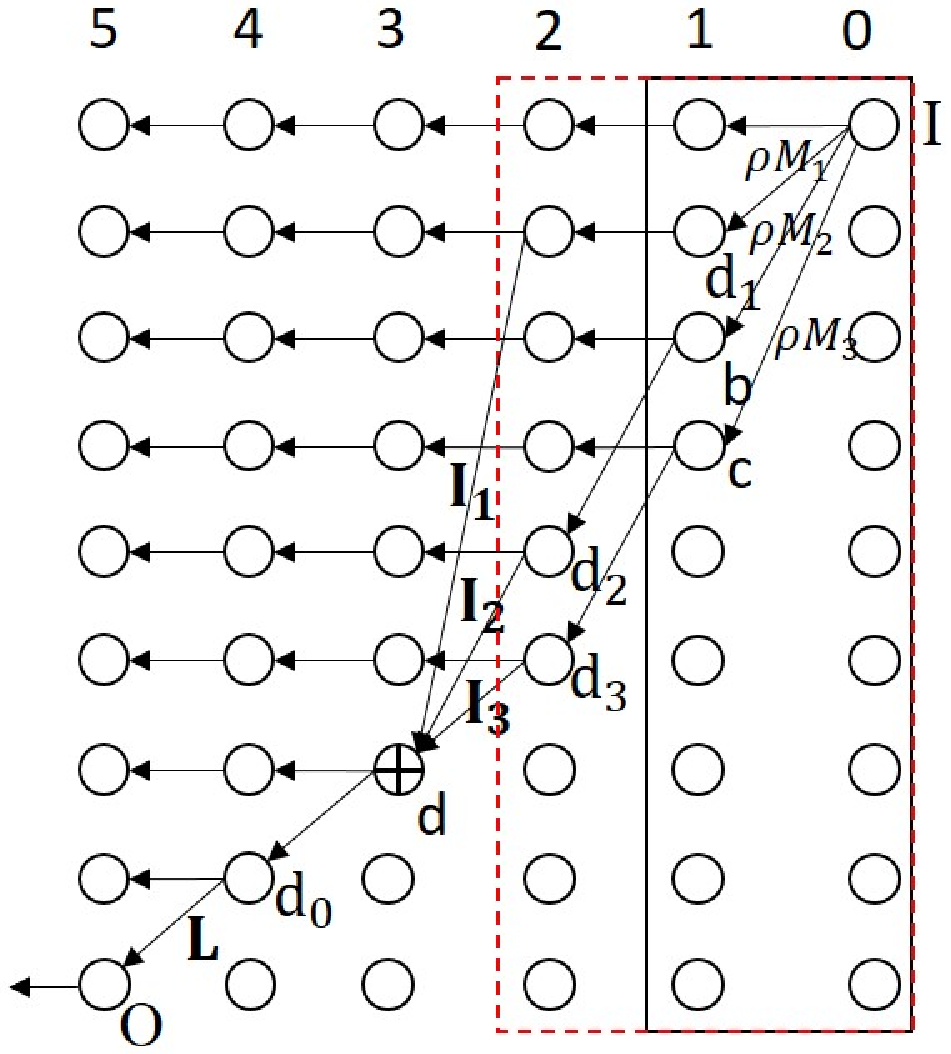}}
\subfigure[]{
\includegraphics[width=0.36\columnwidth]{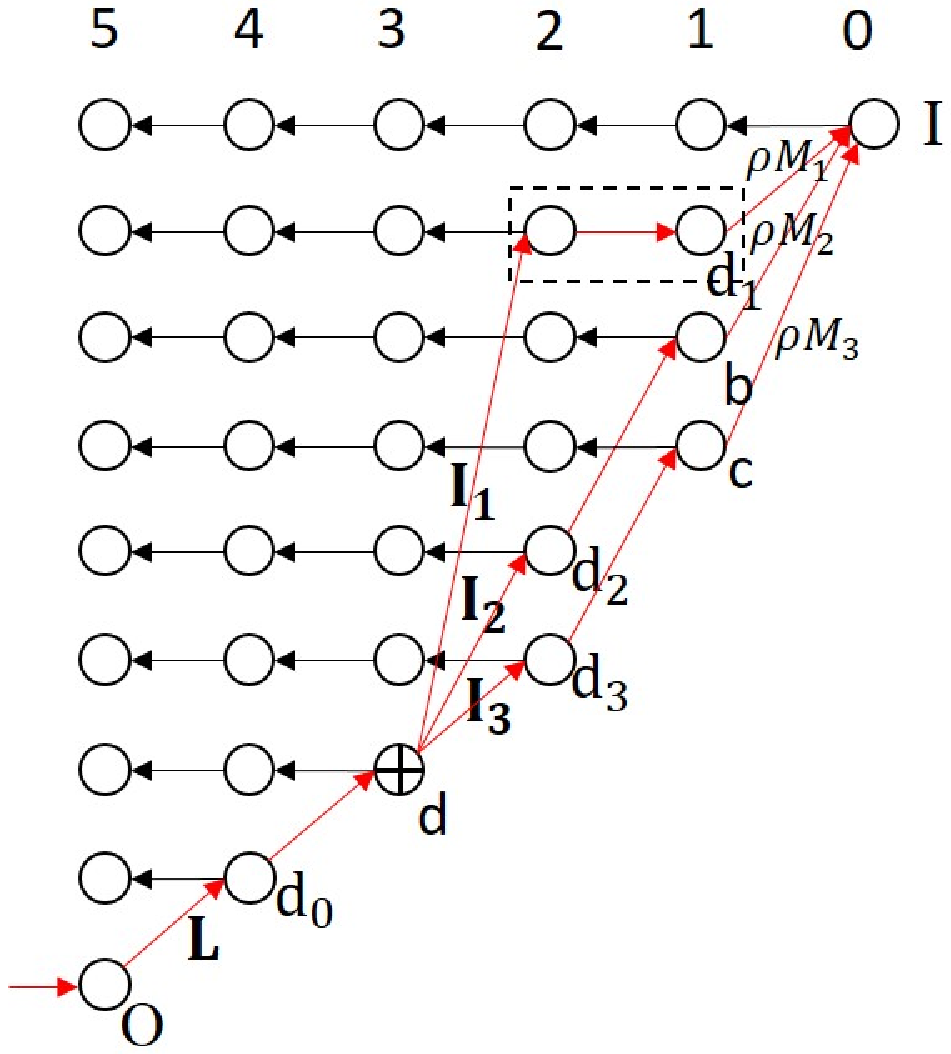}}} 
\caption{
(a) $\mathcal M$ and (b) $\mathcal M^{\oplus}$, where the numbers next to nodes indicate the level of the node and  $\bL$ denotes a linear operation; (c) Graphical representations of $\bC_{\leq n}$ sequences where $n=0, \cdots, 5$ (levels numbered above the columns of nodes). The first row indicates the input node, the last row indicates the output node, and identity operations are not indicated on arcs. Solid box enclosing columns  $0$ and $1$ corresponds to $\bC_{\leq 1}$, while the dashed box in red enclosing columns $0$, $1$, and $2$ corresponds to $\bC_{\leq 2}$; (d)
Back-tracking from output node $O$ to input node $I$ is highlighted in red, and reverse arcs make it possible to obtain $\mathcal M^\oplus$ (isolated nodes are not shown).
} \label{graphC}
\end{figure}

{\bf Remark 1}. \\
One can envision $\bC_{\leq L}$ as an invertible graphical embedding of $\mathcal M^\oplus$ (i.e., $\mathcal M^\oplus$ can be derived from $\bC_{\leq L}$ without ambiguity). This perspective is illustrated by the example in Fig. \ref{graphC}. The graph in Fig. \ref{graphC}(c) depicts how the sequence of $\bC_{\leq n}$ for the graph in Fig. \ref{graphC}(b) is obtained via (\ref{tworelation}), where $\bB_{n+1, n}$ defines the arcs from nodes in column $n$ to nodes in column $n+1$. The sub-graph under columns $0$ to $n$ forms $\bC_{\leq n}$. Fig. \ref{graphC}(c) illustrates the embedding of Fig. \ref{graphC}(b), while Fig. \ref{graphC}(d) shows that $\mathcal M^\oplus$ can be derived from the embedding without ambiguity (as highlighted in red).

\subsubsection{Functions on complete sub-graphs of DAG-DNNs}

A sub-graph is complete if all inputs required to evaluate the function on the sub-graph are contained in it. As such, all input nodes to an addition-node must be included in a sub-graph to make it a complete sub-graph.

%
%

\begin{citeddef} \label{completedef}
(i) An addition-node in $\mathcal M^\oplus$ is said to be complete within a given sub-graph, if and only if all inputs to the node are included in the sub-graph. \\
(ii) The complete sub-graph between nodes $j$ and $i$ includes only the paths between the two nodes, and every addition-node along each path is complete. The complete sub-graph of node $i$ is an abbreviation for the complete sub-graph between input node $I$ and node $i$.
\end{citeddef}

Not all sub-graphs in a DAG-DNN are complete, a complete sub-graph is self-contained and can be isolated to compute a function within the DAG-DNN. Fig. \ref{levelgraph}(d) presents two complete sub-graphs in Fig. \ref{levelgraph}(c).
Clearly, if node $b$ is contained in the complete sub-graph of node $a$, then the complete sub-graph of node $b$ is subsumed in the complete sub-graph of node $a$. The collection of complete sub-graphs is a subset of the sub-graphs. Thus, complete sub-graphs up to level $n$ can be derived from masking out entries in $\bC_{\leq n}$ corresponding to functions on in-complete sub-graphs. Figs. \ref{stacking}(i) and (j) present functions on complete sub-graphs by setting the value zero to entries of $\bI_1$ or $\bI_2$ in $\bC_{\leq 3}$ and $\bC_{\leq 4}$, respectively. Sub-DNNs in Fig. \ref{stacking0}(b) are $\text{sub-DNN}[4,2], \text{sub-DNN}[5,3], \text{sub-DNN}[i,1]$ with $i=2, \cdots, 7$.



\subsection{Function evaluations}

The conclusion drawn by Theorem \ref{universal} pertaining to the $\tilde \cB$-matrix representation of all functions in $\mathcal M$ facilitates operations (\ref{additionalgebra}) and (\ref{productalgebra}). An evaluation involving the representation at input $\bx$ is denoted as follows:   
\begin{align} \label{evaquation}
\bC_{\leq L}(\bx)  =  \bB_{L,L-1} \bC_{\leq L-1}(\bx) = \bB_{L,L-1}\cdots \bB_{1,0} \bC_0(\bx).
\end{align}
where $\bC_{\leq n}(\bz)$ denotes the evaluation of $\bC_{\leq n}$ at input $\bz$ as a vector. If function $f_{\mathcal M^\oplus}[a,c] \in \cB$ is at entry $(a, c)$ of $\bC_{\leq n}$, then the evaluation $f_{\mathcal M^\oplus}[a,c](\bz)$ is a vector at the output of node $a$ with input $\bz$ to node $c$.

In accordance with Definition \ref{def-level-graph} (iii), $m$ denotes the number of nodes in $\mathcal M^\oplus$ and $m_n$ denotes the number of nodes at level $n$. Suppose that nodes are indices from $0$ to $m-1$. In this evaluation, each node is associated with a vector. We denote $d_{i}$ and $d = \sum_{i=0}^{m-1}d_i$ as the dimensions of the vector with node $i$ and with all $m$ nodes, respectively. We also denote $\bC_{\leq n}(\bx)$ as the evaluation at level $n$ for input $\bx \in \R^{d_0}$, where $d_0$ is the input dimension of $\mathcal M^\oplus$.  
The evaluation vector of input $\bx$ at a given level can be derived via recursion. \\
At input level $0$: $\bx^{(0)}\in \R^d=  [\bx^\top, \b0^\top, \cdots, \b0^\top]^\top$ can be obtained via 
\begin{align} \label{inputnode}
\bC_0(\bx) = \begin{bmatrix} \bI &  \b0 & \cdots & \b0\\
 \b0 & \bI & \b0 & \cdots\\
 \vdots &  \vdots  &  \vdots &  \vdots \\
 \b0 & \cdots & \b0 & \bI \end{bmatrix} \begin{bmatrix} \bx \\ \b0 \\ \vdots \\ \b0 \end{bmatrix} = \bx^{(0)}.
 \end{align}
The evaluation at level $n+1$ is a vector in $\R^d$ obtained using (\ref{twolevelgraphs}): 
\begin{align} \label{liftingevaluation}
\bx^{(n+1)} = \bC_{\leq n+1}(\bx) = \bB_{n+1, n}\bx^{(n)} = \bB_{n+1, n}\bC_{\leq n}(\bx) .
\end{align}
This equation demonstrates that the evaluation of $\bC_{\leq n}$ with respect to input $\bx$ can be derived recursively. 
Vector $\bx^{(L)}$ is the output of $\bC_{\leq L}(\bx)$. $\bx^{(n)}$ can be deemed as an $m$-block vector where $\bx^{(n)}  =  [(\bx^{(n)}_0)^\top, \cdots, (\bx^{(n)}_{m-1})^\top]^\top$, wherein $\bx^{(n)}_i \in \R^{d_i}$ is the block associated with node $i$. 
We also let  
$\bx^{(n)}_{\leq n}$ denote the restriction of $\bx^{(n)}$ to the nodes up to level $n$:
\begin{align} \label{restriction}
\bx^{(n)}_{\leq n} = [(\bx^{(n)}_0)^\top, \cdots, (\bx^{(n)}_n)^\top]^\top.
\end{align}
In accordance with (\ref{inputnode}) and (\ref{liftingevaluation}), $\bx^{(n)}$ with nodes at levels above $n$ are zeros and $\bx^{(n+1)}$  and $\bx^{(n)}$ differ only at components corresponding to nodes at level $n+1$. 

\begin{figure}[th]
\centering
\subfigure[]{
\includegraphics[width=0.7\textwidth]{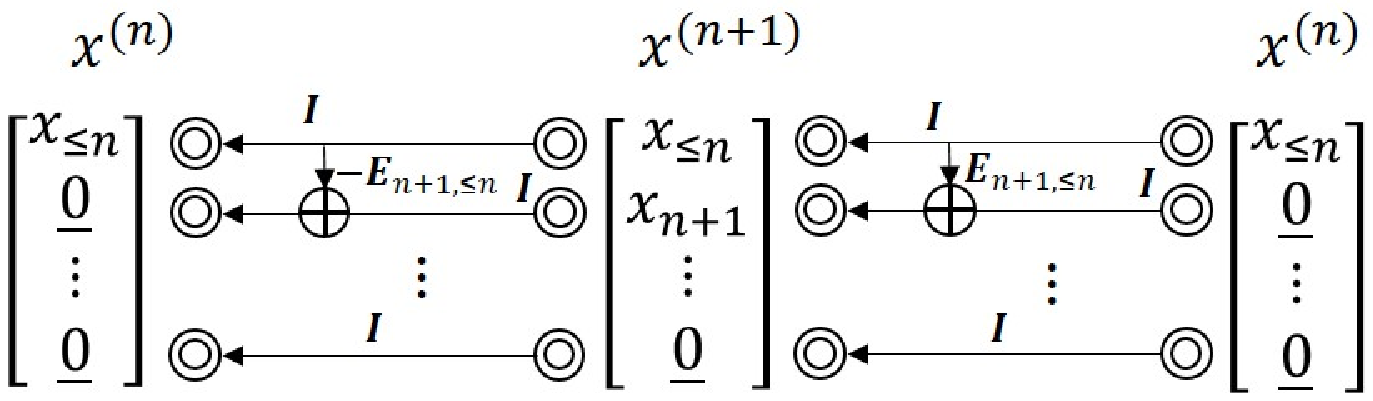}}
\vspace{0.2in}
\subfigure[]
{\includegraphics[width=0.7\textwidth]{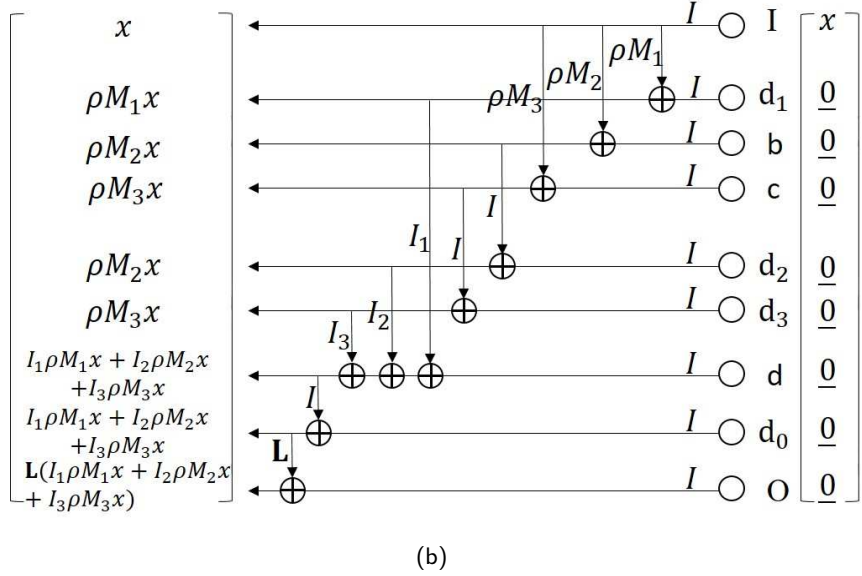}}
\caption{(a) Lifting and inverse lifting structures on level-domain nodes (double-circled nodes) where level-domain nodes in top row include all of the nodes up to level $n$, and the level-domain nodes in second row are at level $n+1$;
 (b) Lifting structure on nodes in Fig. \ref{stacking0}(b) (corresponding name listed in Fig. \ref{graphC}(b) next to nodes). 
} \label{evaluationlift}
\end{figure}

Note that $\bB_{n+1, n}$ is a lower triangular matrix with identity functions at diagonal elements.  According to (\ref{liftingevaluation}), the lower triangle matrix $\bB_{n+1, n}$ presents a lifting scheme by which the all-pair functions up to one level is lifted to the next level in level-graphs. Lower triangle matrix $\bB^{-1}_{n+1, n}$ presents a similar lifting scheme in reverse order. The lifting and inverse lifting structure can be applied to nodes (vertices) and level-domain nodes (i.e., collections of nodes up to a level or at the same level).  
Fig. \ref{evaluationlift}(a) demonstrates the lifting and inverse lifting for evaluations of functions on level-domain nodes. Fig. \ref{evaluationlift}(b) demonstrates the lifting on nodes (vertices) in Fig. \ref{stacking}(b).

{\bf Remark 2.} \\
We have demonstrated that the network invertible to a given network can be derived using the lifting scheme; however, this does not imply that all DAG-DNNs are invertible by which we mean input $\bx$ can be derived from $\by = \mathcal M \bx$ with given $\by$ and $\mathcal M$ because $\by$ is the sub-vector corresponding to the output node in $\bx^{(L)}$. For example, we cannot derive $\bx$ from $\by = \text{(ReLU)} \mathcal M \bx$, due to the fact that ReLU is not an invertible function. Nevertheless, we can obtain the invertible network to $\text{ReLU} \mathcal M$ as
\begin{align*}
\begin{bmatrix} \bI & \b0 \\
-\text{(ReLU)} \mathcal M & \bI 
\end{bmatrix}.
\end{align*}


\begin{figure}[th]
\centering
\subfigure[]{
\includegraphics[width=0.4\textwidth]{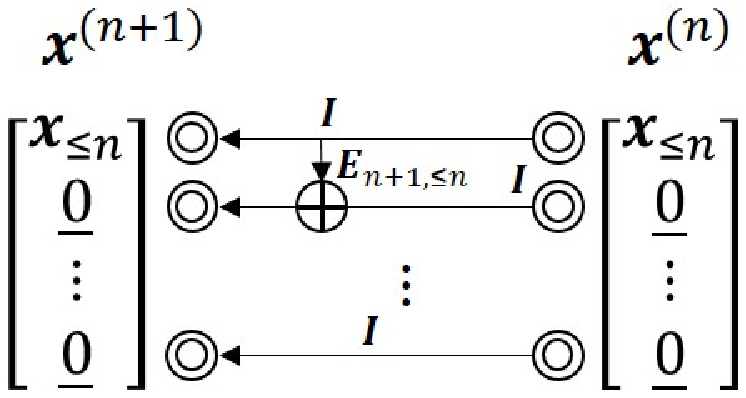}}
\subfigure[]
{
\includegraphics[width=0.4\textwidth]{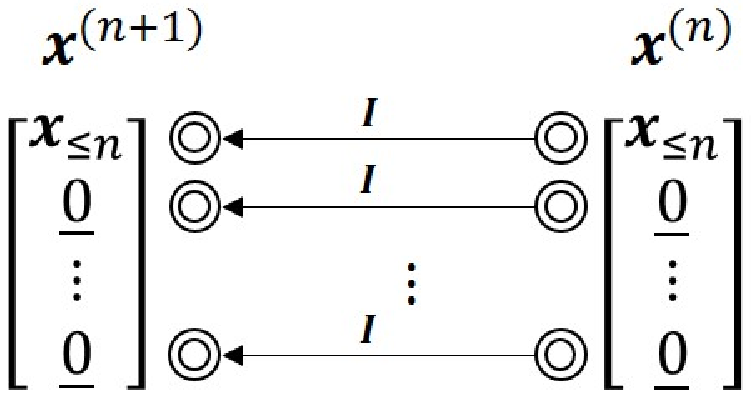}}
\caption{(a) Solutions that satisfy $\bE_{n+1, \leq n} \bx_{\leq n} = \b0$ do not alter the value of $\bx_{n+1}$ at $\bx^{(n+1)}$ (i.e., both $\bx_{n+1}^{(n+1)}$ and $\bx_{n+1}^{(n)}$ are zeros, as shown in the second row); (b) Removing $\bE_{n+1, \leq n}$ from (a) or setting $\bE_{n+1, \leq n} = \b0$ yields the same function evaluation shown in (a).  Note that level-domain nodes in the first rows denote the values of nodes up to level $n$.
} \label{leveldomainprun}
\end{figure}

\section{Network compression with structural pruning} \label{sec:compression}

The theory of lower-triangle factorization has a connection to structural pruning (involving the removal of entire neurons and their connections), as it can be used to approximate functions defined for every sub-graph of a DAG-DNN using a sequence of lifting moduli. 
In the following, we outline the pruning conditions for structures in a network that achieves a weak version of the lottery ticket hypothesis, where a dense neural network contains a sub-network that can be initialized and trained in isolation with ``training accuracy" on par with that of the original network using the same number of iterations or fewer. 
In accordance with the lifting structure in Fig. \ref{leveldomainprun}(a), we obtain the following:
\begin{align} \label{pruning}
\bx_{n+1}^{(n+1)} = \bE_{n+1, \leq n} \bx_{\leq n}^{(n)} + \bx_{n+1}^{(n)} =\bE_{n+1, \leq n} \bx_{\leq n}^{(n)},
\end{align}
where $\bx_{\leq n}^{(n)}$, in (\ref{restriction}), represents the value of the level-domain nodes up  to level $n$, and $\bx_{n+1}$ represents the value of the level-domain node at level $n+1$ . We know that the second equality in (\ref{pruning}) can be obtained, based on the fact that the values of nodes at level above $n$ are zeros in $\bx^{(n)}$. Thus, if $ \bE_{n+1, \leq n} \bx_{\leq n}^{(n)} = \b0$,  then $\bx_{n+1}^{(n+1)} = \bx_{n+1}^{(n)} = \b0$. As such, all components in $\bE_{n+1, \leq n}$ can be removed (equivalent to removing connections from nodes at level $\leq n$ to nodes at level $n+1$; see Fig. \ref{leveldomainprun}(b)) without affecting the function evaluation with respect to the input $\bx$. 

Given $N$ training data $\{(\bu_i, \bv_i)\}$, suppose that the loss $L$ in training $\mathcal M_0^\oplus$ with parameter $\theta_0$ is as follows:
\begin{align} \label{trainingloss}
L(\theta_0) = \frac{1}{N} \sum_i  l(\bv_i - \mathcal M^\oplus_0(\theta_0) \bu_i) + R(\theta_0),
\end{align}
where $l$ measures the prediction fidelity (accuracy) between $\bv_i$ and $\mathcal M_0^\oplus(\theta_0) \bu_i$.  The first term of (\ref{trainingloss}) is the expected training loss. $R(\theta_0)$ is a regulatory function in which any subset $\theta$ of $\theta_0$ yields $R(|\theta|) \leq R(|\theta_0|)$, where $| \cdot |$ indicates the size.
Suppose that $\mathcal M_0^\oplus$ can be trained using descent algorithm $\cA$ to parameter $\theta_0^*$ at cost $L(\theta_0^*)$, at which the expected training loss $\epsilon_0^*$ is minimized. We respectively denote $\mathcal M^\oplus_0(\theta(t))$ and $L(\theta(t))$ as the trained machine and training loss at time $t$.

\begin{citedthm} \label{weakhypo}
Suppose the assumptions pertaining to $\mathcal M_0^\oplus$ in (\ref{trainingloss}) and algorithm $\cA$ for training data $\{(\bu_i, \bv_i)\}_{i=1}^N$ hold true. In accordance with  (\ref{liftingevaluation}) and (\ref{restriction}) at time $t$, we let  
\begin{align*} 
\bX^{(n)}(t) = [ \bu_1^{(n)}(t), \cdots, \bu_N^{(n)}(t)]
\end{align*} and 
\begin{align*}
\bX^{(n)}_{\leq n}(t) = [ (\bu_1^{(n)})_{\leq n}(t), \cdots, (\bu_N^{(n)})_{\leq n}(t)].
\end{align*}
We also denote $[\bE_{n+1, \leq n}]_i$ as the $i$-th row of $\bE_{n+1, \leq n}$ and let 
\begin{align}
Z_{n+1, \leq n}(t) =\{ i \;|  [\bE_{n+1, \leq n}(t)]_i \bX^{(n)}_{\leq n}(t)  = \b0\};
\end{align}
i.e., the collection of nodes at level $n+1$ with values of zero in corresponding rows of 
$\bE_{n+1, \leq n}(t) \bX^{(n)}_{\leq n}(t)$.  
Suppose that $\mathcal M_1^\oplus$ is the sub-network of $\mathcal M_0^\oplus$ after removing the components in $\bE_{n+1, \leq n}$ corresponding to nodes in $Z_{n+1, \leq n}(t)$. The parameter of $\mathcal M_1^\oplus$ is $\theta_1$. 
If  
\begin{align} \label{pruncon}
Z_{n+1, \leq n}(t) \neq \emptyset,
\end{align}
for some $n \in [0,  L-1]$ at time $t$, and if 
 \begin{align} \label{pruncon1}
\begin{cases}
L(\theta_1(t)) \leq L(\theta^*) \\
R(|\theta_0|) - R(|\theta_1|) \leq c \epsilon_0^*,
\end{cases}
\end{align}
for some constant $c$, depending on $n$ and $t$, 
then the weak version of the lotus ticket hypothesis is asserted using winning ticket $[\mathcal M_1^\oplus, \theta_1(t)]$.
\end{citedthm}
\proof
Let $i \in Z_{n+1, \leq n}(t)$. 
Since $[\bE_{n+1, \leq n}(t)]_i \bX^{(n)}_{\leq n}(t) = \b0$, the removal of components in $[\bE_{n+1, \leq n}(t)]_i$ from $\mathcal M_0^\oplus(t)$ would not alter the expected empirical loss at time $t$.
Because $R(|\theta_1|) \leq R(|\theta_0|)$ ($\theta_1 \subseteq \theta_0$),  we obtain  $L(\theta_1(t)) \leq L(\theta_0(t))$. Since $L(\theta_1(t)) \leq L(\theta^*)$ and $\cA$ is a descent algorithm, applying $\cA$ to train $\mathcal M_1^\oplus$ with parameters initialized at $\theta_1(t)$ leads to a decrease in loss with $L(\theta_1(t^+)) \leq L(\theta^*)$ for $t^+ \geq t$.  Substituting (\ref{trainingloss}) into $L(\theta_1(t^+)) \leq L(\theta^*)$ and re-arranging the term $ R(|\theta_1|)$ yields the following:
\begin{align}
\frac{1}{N} \sum_i  l(\bv_i - \mathcal M_1^\oplus (\theta_1(t^+)) \bu_i) \leq \epsilon_0^* + R(|\theta_0|) - R(|\theta_1|) \leq (c+1) \epsilon_0^*.
\end{align}
Thus, when training $\mathcal M_1^\oplus$ using parameters initialized at $\theta_1(t)$, it is possible to match the expected training loss $\epsilon_0^*$ (best training performance) of the original network up to a factor of $(c+1)$ for any iteration $t^+ -t \geq 0$.

\qed

Conditions (\ref{pruncon}) and (\ref{pruncon1}) are sufficient to allow rewinding, which involves the rewinding of weight coefficients backwards in training the original network, while using the coefficients for initialization in training a sub-network. With minimal modification, we can apply the above theorem to this technique to obtain the following corollary.
\begin{citedcor}
Suppose that the assumptions pertaining to Theorem \ref{weakhypo} hold true. Further suppose that $\mathcal M_0^\oplus$ can be trained using algorithm $\cA$ to match parameter $\theta_0(t_0)$ at $t_0$, where the expected training loss is $\epsilon_0(t_0)$. We can then obtain the winning ticket $[\mathcal M_1^\oplus, \theta_1(t)]$, the derivation of which is based on the assumption that $n$ and $c$ exist at $t < t_0$, such that
\begin{align} \label{pruncon2}
\begin{cases}
Z_{n+1, \leq n}(t) \neq \emptyset \\
L(\theta_1(t)) \leq L(\theta_0(t)) \\
R(|\theta_0|) - R(|\theta_1|) \leq c \epsilon_0(t_0).
\end{cases}
\end{align}
The ticket can then be used to train $\mathcal M_1^\oplus(\theta_1(t^+))$ using $\cA$ to match $\epsilon_0(t_0)$ up to a factor of $(c+1)$ for $t \leq t^+ \leq t_0$.
\end{citedcor}
This allows us to rewind the training coefficients to $\mathcal M_0^\oplus$ at $t_0$. If (\ref{pruncon2}) is satisfied at $t \in [0, t_0]$ for some $n$ and $c$, then it is possible to start training $\mathcal M_1^\oplus$ using weight coefficients initialized at $\theta_1(t)$ to $t_0$. We know that the training accuracy of the sub-network is ensured, as the the corollary holds for $\epsilon_0(t_0)$ up to a factor of $c+1$ for any number of iterations $t^+ \leq t_0$. It is possible to apply the above procedure repeatedly via multiple rewinding loops to yield networks of ever-decreasing size without compromising the training accuracy of the original network.

{\bf Remark 3}. The regulatory function is defined according to the size of network parameters. Note that this is  not unusual, considering that when using a classification tree, the number of terminal nodes is regularized to find a tree that balances training accuracy and tree-size \cite{breiman2017classification}. When performing network pruning, treating network-size as a regulatory function results in a similar trade-off between learning accuracy and network complexity.

\section{Conclusions} \label{sec:conclusions}

This paper uses DAG-DNN graphical representation to provide a universal representation of all functions defined for every sub-graph of a given DNN. This paper also outlines the theory of lower-triangle factorization, in which the representations of functions are expressed as multiplications of lower triangular matrices, each of which characterizes functions over sub-graphs to nodes at a specified level. The lifting structure associated with the lower triangular matrices allows systematic structural pruning, regardless of the underlying DNN architecture.
We demonstrate that it is theoretically possible to derive a sub-network that can be initialized and trained in isolation with training accuracy on part with that of the original network using the same number of iterations or fewer. 
We expect that other properties of DNNs could be derived by leveraging host analysis tools for graphs. \\

{\bf{Acknowledgements}}: Wen-Liang Hwang would like to express his gratitude to Mr. Shih-Shuo Tung at Institute of Information Science, Academia Sinica, Mr. Ming-Yu Chung, at Department of Electrical Engineering, Taiwan University, and Dr. Pin-Yu Chen at IBM Research AI for assistances and valuable comments on the presentation of this paper.

\bibliographystyle{ieeetr}
\bibliography{DeepRef}

\begin{thebibliography}{10}

\bibitem{radhakrishnan2023wide}
A.~Radhakrishnan, M.~Belkin, and C.~Uhler, ``Wide and deep neural networks
  achieve consistency for classification,'' {\em Proceedings of the National
  Academy of Sciences}, vol.~120, no.~14, p.~e2208779120, 2023.

\bibitem{Cybenko1989}
G.~Cybenko, ``Approximation by superpositions of a sigmoidal function,'' {\em
  Mathematics of Control, Signals and Systems}, vol.~2, no.~4, pp.~303--314,
  1989.

\bibitem{Hornik91}
K.~Hornik, ``Approximation capabilities of multilayer feedforward networks,''
  {\em Neural Networks}, vol.~4, no.~2, pp.~251--257, 1991.

\bibitem{heinecke2020refinement}
A.~Heinecke, J.~Ho, and W.-L. Hwang, ``Refinement and universal approximation
  via sparsely connected relu convolution nets,'' {\em IEEE Signal Processing
  Letters}, vol.~27, pp.~1175--1179, 2020.

\bibitem{neal2012bayesian}
R.~M. Neal, {\em Bayesian learning for neural networks}, vol.~118.
\newblock Springer Science \& Business Media, 2012.

\bibitem{williams1998prediction}
C.~K. Williams, ``Prediction with gaussian processes: From linear regression to
  linear prediction and beyond,'' {\em Learning in graphical models},
  pp.~599--621, 1998.

\bibitem{lee2017deep}
J.~Lee, Y.~Bahri, R.~Novak, S.~S. Schoenholz, J.~Pennington, and
  J.~Sohl-Dickstein, ``Deep neural networks as gaussian processes,'' {\em arXiv
  preprint arXiv:1711.00165}, 2017.

\bibitem{jacot2018neural}
A.~Jacot, F.~Gabriel, and C.~Hongler, ``Neural tangent kernel: Convergence and
  generalization in neural networks,'' {\em Advances in neural information
  processing systems}, vol.~31, 2018.

\bibitem{BaraniukPowerDiagramSubdiv}
R.~Balestriero, R.~Cosentino, B.~Aazhang, and R.~Baraniuk, ``The geometry of
  deep networks: Power diagram subdivision,'' {\em Advances Neural Inf.
  Process. Syst.}, pp.~15806--15815, 2019.

\bibitem{hwang2019rectifying}
W.-L. Hwang and A.~Heinecke, ``Un-rectifying non-linear networks for signal
  representation,'' {\em IEEE Transactions on Signal Processing}, vol.~68,
  pp.~196--210, 2019.

\bibitem{hwang2022analysis}
W.-L. Hwang and S.-S. Tung, ``Analysis of function approximation and stability
  of general dnns in directed acyclic graphs using un-rectifying analysis,''
  {\em arXiv preprint arXiv:2206.05997}, 2022.

\bibitem{frankle2018lottery}
J.~Frankle and M.~Carbin, ``The lottery ticket hypothesis: Finding sparse,
  trainable neural networks,'' {\em arXiv preprint arXiv:1803.03635}, 2018.

\bibitem{malach2020proving}
E.~Malach, G.~Yehudai, S.~Shalev-Schwartz, and O.~Shamir, ``Proving the lottery
  ticket hypothesis: Pruning is all you need,'' in {\em International
  Conference on Machine Learning}, pp.~6682--6691, PMLR, 2020.

\bibitem{zhang2021lottery}
S.~Zhang, M.~Wang, S.~Liu, P.-Y. Chen, and J.~Xiong, ``Why lottery ticket wins?
  a theoretical perspective of sample complexity on sparse neural networks,''
  {\em Advances in Neural Information Processing Systems}, vol.~34,
  pp.~2707--2720, 2021.

\bibitem{arora2016understanding}
R.~Arora, A.~Basu, P.~Mianjy, and A.~Mukherjee, ``Understanding deep neural
  networks with rectified linear units,'' {\em arXiv preprint
  arXiv:1611.01491}, 2016.

\bibitem{renda2020comparing}
A.~Renda, J.~Frankle, and M.~Carbin, ``Comparing rewinding and fine-tuning in
  neural network pruning,'' {\em arXiv preprint arXiv:2003.02389}, 2020.

\bibitem{zhou2019deconstructing}
H.~Zhou, J.~Lan, R.~Liu, and J.~Yosinski, ``Deconstructing lottery tickets:
  Zeros, signs, and the supermask,'' {\em Advances in neural information
  processing systems}, vol.~32, 2019.

\bibitem{breiman2017classification}
L.~Breiman, {\em Classification and regression trees}.
\newblock Routledge, 2017.

\end{thebibliography}

\end{document}